\documentclass{article} % For LaTeX2e
\usepackage{iclr2027_conference,times}

\usepackage{amsmath,amsfonts,bm}

\def\eqref#1{equation~\ref{#1}}

\def\1{\bm{1}}

\DeclareMathAlphabet{\mathsfit}{\encodingdefault}{\sfdefault}{m}{sl}
\SetMathAlphabet{\mathsfit}{bold}{\encodingdefault}{\sfdefault}{bx}{n}

\usepackage{hyperref}
\usepackage{url}
\usepackage{booktabs}
\usepackage{makecell}
\usepackage{multirow}
\usepackage{graphicx}
\usepackage{amssymb}
\usepackage{wrapfig}
\usepackage{xcolor}
\usepackage{colortbl}
\usepackage{tabularx}
\usepackage{listings}

\usepackage{capt-of}
\usepackage{pgfplots}
\pgfplotsset{compat=1.18}
\definecolor{LEAblMemory}{HTML}{F4B8C5}
\definecolor{LEAblRetrieval}{HTML}{F7D79A}
\definecolor{LEAblLongEmo}{HTML}{B9B3E8}

\definecolor{LEGroupBg}{HTML}{EDF1F5}
\definecolor{LEAccent}{HTML}{2D465C}

\newcommand{\LEyes}{\ensuremath{\checkmark}}
\newcommand{\LEno}{\textcolor{black!50}{\textendash}}

\title{LongEmo: Towards Emotion Understanding and Reasoning in Long Videos}

\author{%
\parbox{\dimexpr\textwidth-2\tabcolsep\relax}{\raggedright
\mbox{Shuo Zhang$^{1,*}$}, \mbox{Yifan Zhou$^{2,*}$}, \mbox{Han Wang$^{3,*}$}, \mbox{Jinsong Zhang$^{4,*}$}, \mbox{Jingyu Li$^{5,\dagger}$},\\
\mbox{Hongbing Li$^{1}$}, \mbox{Zhejun Zhang$^{1}$}, \mbox{Chengyi Zhao$^{6}$}, \mbox{Yuquan Hao$^{1}$}, \mbox{Yitong Liu$^{1}$},\\
\mbox{Jiyin Li$^{1}$}, \mbox{Ruiqi Tang$^{1}$}, \mbox{Zixuan Lin$^{1}$}, \mbox{Yi Luo$^{1}$}, \mbox{Xurui Zhang$^{7}$}, \mbox{Ronghao Chen$^{8,\dagger}$},\\
\mbox{Huacan Wang$^{9,\dagger}$}, \mbox{Lei Li$^{1,\dagger}$}\\[0.5em]
\normalfont\small
\mbox{$^{1}$BUPT}\quad \mbox{$^{2}$SJTU}\quad \mbox{$^{3}$THU}\quad
\mbox{$^{4}$HIT}\quad \mbox{$^{5}$USTC}\quad \mbox{$^{6}$BNU}\quad
\mbox{$^{7}$CUFE}\quad \mbox{$^{8}$PKU}\quad \mbox{$^{9}$UCAS}
}}

\iclrfinalcopy % Display authors in this arXiv version.
\begin{document}

\maketitle
\begingroup
\renewcommand{\thefootnote}{\fnsymbol{footnote}}
\footnotetext[1]{Equal contribution.}
\footnotetext[2]{Corresponding authors. Email: \href{mailto:shuoz@bupt.edu.cn}{\texttt{shuoz@bupt.edu.cn}}.}
\endgroup
\lhead{} % No conference publication claim in the arXiv version.

\begin{abstract}
While recent Multimodal Large Language Models (MLLMs) have shown promise in affective computing, their reasoning capabilities are largely confined to short video clips with limited interactions. However, real-world emotions are not merely isolated instantaneous reactions but dynamic and cumulative processes deeply shaped by past experiences and ongoing events. To bridge this gap, we introduce LongEmoBench, a benchmark dedicated to emotion understanding and reasoning in long videos. It assesses progressive capabilities scaling from continuous scene interactions to complex episodic developments. Furthermore, we propose LongEmo, a novel memory-augmented agentic framework designed to tackle the immense challenges of long-range affective reasoning. LongEmo processes continuous video streams to construct an Event Memory Graph, explicitly modeling long-range dependencies and capturing emotional dynamics across discrete events. Given a question, the agent retrieves a query-relevant event stream from the graph, iteratively integrating multimodal memories and relational dependencies to deduce the final answer. Extensive evaluations of 17 representative methods reveal that they struggle significantly with emotion understanding and reasoning in long videos. In contrast, LongEmo achieves state-of-the-art performance, demonstrating the efficacy of its event-centric memory architecture. 
\end{abstract}

\section{Introduction}

Understanding human emotions is a cornerstone for developing human-centric AI. Recent advances in Multimodal Large Language Models (MLLMs)~\citep{cheng2024emotionllama,lian2025affectgpt,zhao2025r1omni} have spurred substantial progress in affective computing. Beyond recognizing isolated expressions~\citep{lian2025ovmer} from facial muscle movements, vocal acoustics, and lexical semantics, modern models can explain the immediate causes of emotional reactions~\citep{zhang2025videmo}, and infer affective shifts within short local dialogues~\citep{hu2026emotrans}.

However, existing methods largely confine reasoning to extremely short, isolated video clips. In real-world scenarios, emotion is rarely a simple slice of a transient state, but a dynamic process accumulated over time~\citep{scherer2009emotions}. A character's present feelings are deeply influenced by past experiences, may be deliberately masked by superficial behavior, and are continuously redefined as events unfold. Accurately inferring such complex states requires traversing the temporal dimension to integrate multimodal clues scattered across long-range contexts. Existing affective benchmarks~\citep{lian2025mer2025,hu2026emobenchm,zhang2026mmeemotion} almost exclusively remain at the level of short clips, single scenes, or single conversations, where the emotional dynamics they capture are restricted to, at most, immediate expressive reactions. This naturally raises the question: \textit{are current MLLMs and video agents capable of performing long-range emotion understanding and reasoning across extended video narratives?}

Although general long-video benchmarks~\citep{fu2025videomme,wang2025lvbench,yang2026egolife} successfully scale inputs to hour-long durations, they primarily assess physical realities and factual plot points rather than characters' internal states, leaving long-term emotion understanding and reasoning largely unaddressed. To bridge this gap, we introduce \textbf{LongEmoBench}, a comprehensive benchmark dedicated to emotion understanding and reasoning in long videos. It comprises 1,975 high-quality QA pairs across approximately 70 hours of video footage, with individual videos spanning from short contextual scenes to nearly hour-scale continuous narratives, far exceeding the temporal span of existing benchmarks (See Table \ref{tab:benchmark_comparison}). We systematically decompose the evaluation into two complementary granularities: Scene-level and Episode-level. Scene-level assesses intra-scene affective comprehension, evaluating standard MLLMs within limited context windows. Episode-level specifically challenges long-context models and advanced video agents, requiring them to synthesize temporally distant evidence for emotion tracking and complex reasoning.

To tackle the immense challenges posed by long-term emotion reasoning, we propose \textbf{LongEmo}, a novel memory-augmented agentic framework. Motivated by the observation that a video fundamentally unfolds as a dynamic event stream within a persistent world, we formulate long-term emotion reasoning as modeling affective dynamics across discrete events. Accordingly, LongEmo distills the untrimmed video into structured event memories, encapsulating characters' emotional states and supporting multimodal evidence. To capture long-range dependencies, we construct a global Event Memory Graph by interconnecting these event memories based on the temporal and causal logic driving emotional dynamics. Given a user query, LongEmo retrieves a relevant event stream from the graph to establish a character-centric emotional context. Subsequently, the reasoning agent executes a progressive access loop, seamlessly integrating multimodal memories within individual events and relational dependencies across the event stream to logically deduce the final answer.

Using LongEmoBench, we evaluate 17 approaches on LongEmoBench, including 9 standard MLLMs and 8 long-video understanding methods (long-video models and agent-based methods). Evaluations reveal that while standard MLLMs capture immediate emotional cues, they struggle with complex causal reasoning and subtle state transitions. Furthermore, although long-video methods handle general tasks adequately, they fall short in emotion reasoning, failing to model long-range emotional dynamics. In contrast, LongEmo achieves state-of-the-art (SOTA) performance, validating the effectiveness of its event-centric memory architecture.

\section{Related Work}

\begin{table}[t]
    \centering
    \caption{
        Comparison of LongEmoBench with existing affective video benchmarks.
        \textbf{Avg. Duration} denotes the average duration of the video provided
for each question, and \textbf{Total Hours} denotes the total
duration of annotated videos.
    }
    \label{tab:benchmark_comparison}

    \vspace{4pt}
    
    \small
    \setlength{\tabcolsep}{3pt}
    \setlength{\arrayrulewidth}{0.4pt}
    \renewcommand{\arraystretch}{1.22}
    \setlength{\aboverulesep}{0pt}
    \setlength{\belowrulesep}{0pt}

    \begin{tabular*}{\linewidth}{
        @{}l
        @{\hspace{6pt}}|@{\hspace{6pt}}
        c
        @{\extracolsep{\fill}}cccc@{}
    }
        \toprule

        \textbf{Benchmark}
        & \textbf{\# Tasks}
        & \textbf{\# Videos}
        & \textbf{\# QA}
        & \textbf{Avg. Duration}
        & \textbf{Total Hours} \\

        \midrule

        OV-MER~\citep{lian2025ovmer}
        & 1 & 332 & 332
        & 4.0\,s & 0.4\,h \\

        EmoTrans~\citep{hu2026emotrans}
        & 4 & 1,000 & 3,274
        & 5.5\,s & 1.5\,h \\

        MtMeUR~\citep{hu2025mtmeur}
        & 5 & 1,451 & 5,101
        & 17.3\,s & 6.8\,h \\

        MME-Emotion~\citep{zhang2026mmeemotion}
        & 8 & 6,500 & 6,500
        & 4.8\,s & 8.6\,h \\

        EmoBench-M~\citep{hu2026emobenchm}
        & 13 & 5,642 & 5,646
        & 7.3\,s & 11.9\,h \\

        \midrule

        \textbf{LongEmoBench (Scene)} 
        & 5 & 1,213 & 1,417
        & \textbf{1.3\,min} & \textbf{23.9\,h} \\

        \textbf{LongEmoBench (Episode)} 
        & 3 & 141 & 558
        & \textbf{20.0\,min} & \textbf{45.9\,h} \\

        \bottomrule
    \end{tabular*}
\end{table}

\textbf{Multimodal Emotion Benchmarks.}
The rapid advancement of MLLMs has fundamentally reshaped benchmarking in affective computing. This evaluation paradigm has continuously evolved, progressing from early closed-set classification~\citep{lian2024merbench,hu2026emobenchm} to open-vocabulary description~\citep{lian2024mer2024,lian2025ovmer}, and extending to causal explanation~\citep{lian2025affectgpt,zhang2026mmeemotion}. To further challenge modern models, recent pioneering works have expanded into more complex interactive scenarios, introducing benchmarks that evaluate emotion transitions~\citep{hu2026emotrans}, multi-turn and multi-party affective dialogues~\citep{hu2025mtmeur,sasu-etal-2025-akan}, and even model robustness against emotional hallucinations~\citep{xing2025emotionhallucer}. However, the video inputs in these studies remain restricted to brief scenes or single interactions, providing only immediate context and thus failing to evaluate a model's ability to capture long-range dependencies and perform complex affective reasoning. In contrast, LongEmoBench transcends these temporal limitations, connecting fine-grained audiovisual cues with extended temporal contexts to push the boundaries of affective computing into the true long-video regime.

\begin{figure}[htbp]
    \centering
    \includegraphics[width=0.85\linewidth]{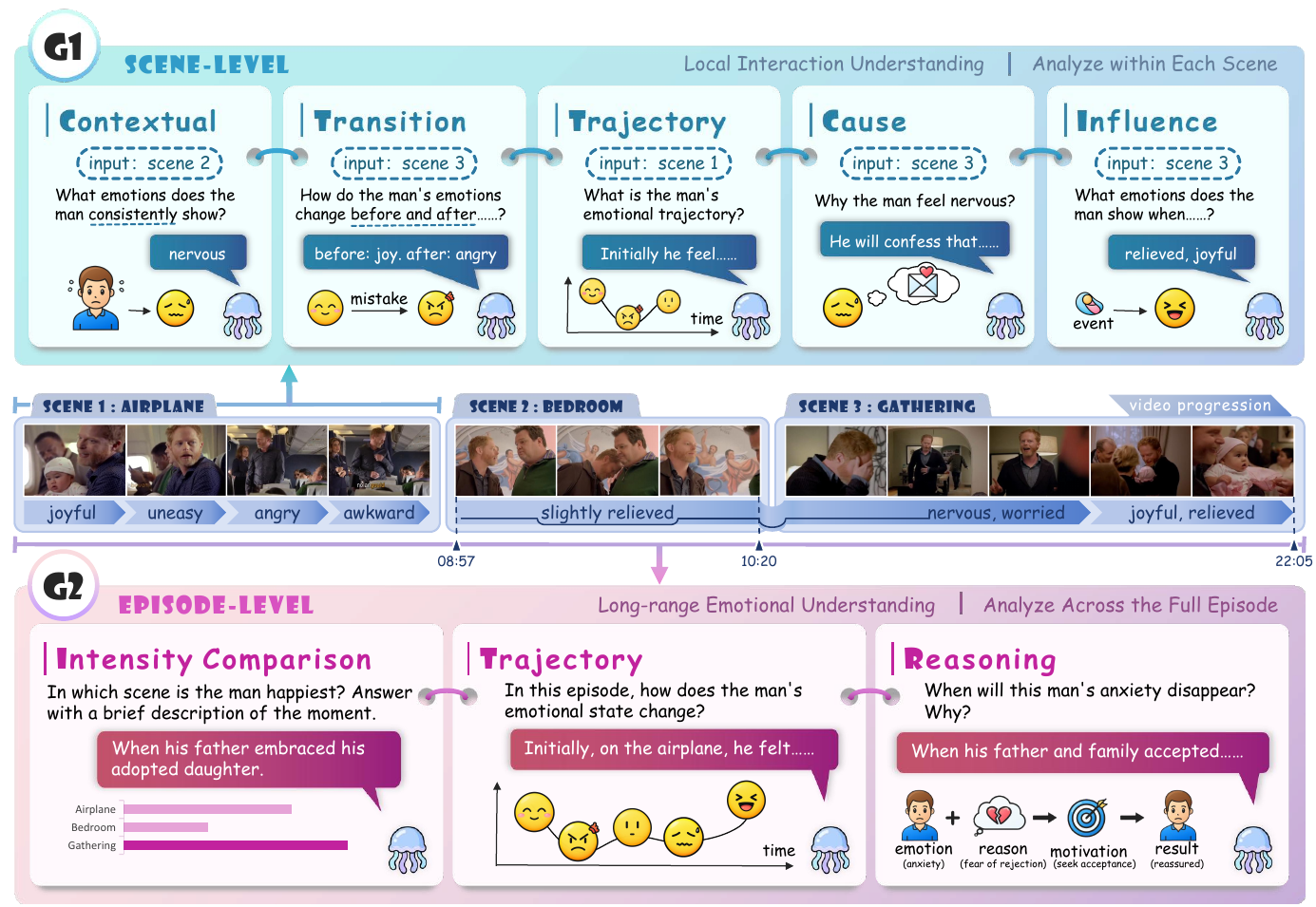}
    \caption{\textbf{Overview of our LongEmoBench.} The G1 (Scene-Level) includes five tasks for localized emotional interactions. The G2 (Episode-Level) includes three tasks for long-range emotional reasoning. A timeline illustrates the temporal scope distinguishing these two granularities.}
    \label{fig:overview}
\end{figure}

\textbf{Long-Video Understanding Methods.}
To process hour-level audiovisual sequences, recent approaches typically tackle long-context bottlenecks through two primary paradigms: agentic frameworks and architectural optimizations. The first paradigm circumvents context constraints by formulating temporal modeling as an active, agent-driven exploration process. Memory-augmented systems like M3-agent~\citep{long2025m3agent}, WorldMM~\citep{yeo2026worldmm}, and Light-Omni~\citep{nie2026lightomni} construct explicit long-term memory banks to iteratively index and recall historical semantic events. Alternatively, autonomous vision agents such as PyVision-RL~\citep{zhao2026pyvisionrl} bypass exhaustive video processing by selectively sampling temporal context strictly on demand. The second paradigm mitigates computational overhead directly at the structural level. Token compression strategies~\citep{cao2026omnifocus,chen2026avoc} apply query-guided or retrieval-inspired algorithms to drastically reduce spatial-temporal redundancy, whereas streaming architectures~\citep{xiao2024streamingllms,lin2026speakwatching} process inputs sequentially to maintain stable long-term reasoning. However, while these methods excel at general long-video reasoning, they risk compromising affective computing by disrupting continuous emotional trajectories, discarding fine-grained cues, and losing distant emotional anchors. To bridge this gap, we introduce LongEmoBench as a vital testbed to evaluate whether these architectures can overcome these bottlenecks and truly master long-term affective reasoning. Concurrently, we propose a novel event-centric memory method explicitly designed to address these limitations and preserve continuous emotional dynamics.

\section{LongEmoBench}

Figure~\ref{fig:overview} presents an overview of LongEmoBench. It is designed to systematically evaluate whether current MLLMs and agentic frameworks can perform precise emotion understanding and reasoning across extended video contexts. Below, we introduce the benchmark's foundational design and evaluation protocol, while extended details are provided in Appendix.

\begin{figure}[h]
    \centering
    \includegraphics[width=0.85\linewidth]{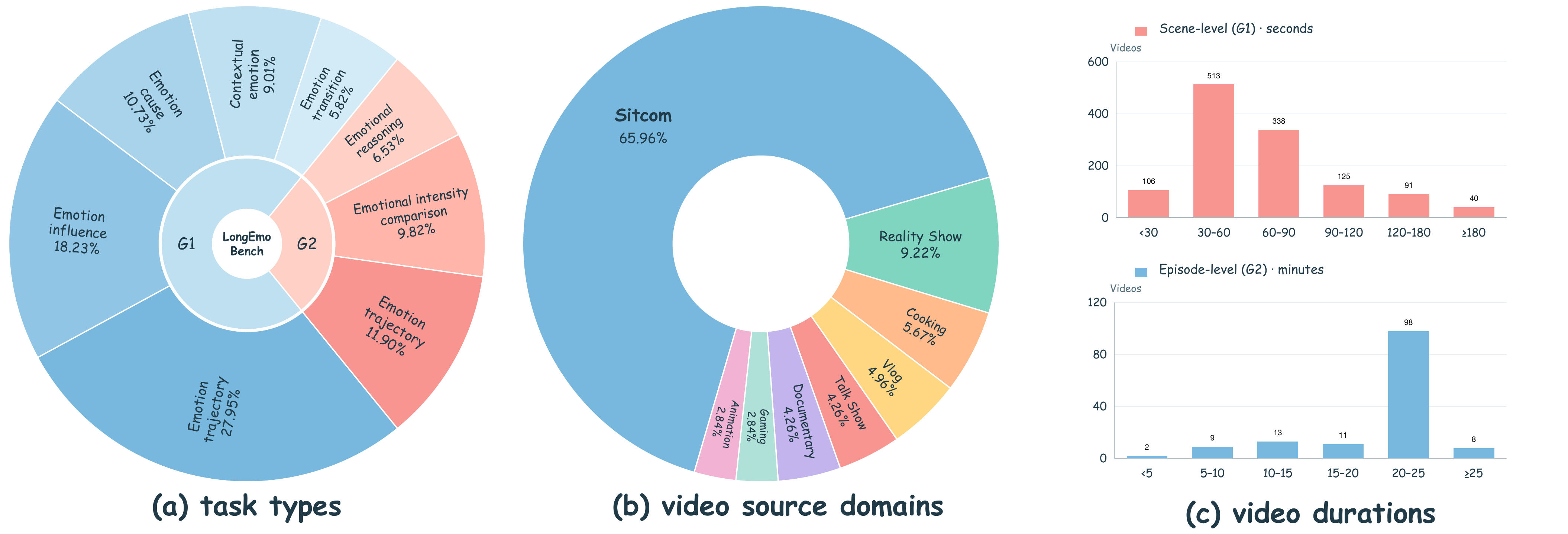}
    \caption{\textbf{Statistical overview of LongEmoBench.} (a) Task type distribution. (b) Video source domain distribution. (c) Video duration distributions for scene and episode levels. }
    \label{fig:statistics}
\end{figure}

\subsection{Task Taxonomy}

LongEmoBench structures its evaluation through a hierarchical task taxonomy grounded in narrative granularity. We organize this taxonomy into two primary tiers, Scene-Level (G1) and Episode-Level (G2), to measure progressive reasoning capabilities that scale from contiguous interpersonal interactions to comprehensive cross-scene narratives. Examples for each task are provided in Appendix~\ref{app:examples}.

\textbf{Scene-Level.} This tier models the complete lifecycle of emotional events within contiguous interactive scenes, mirroring how humans progressively perceive, track, and interpret affective dynamics. Rather than isolated categories, we design five interdependent tasks to provide comprehensive coverage of this cognitive process. \textit{Contextual Emotion} serves as the affective baseline, identifying persistent individual states or collective atmospheres. To track temporal evolution, \textit{Emotion Transition} captures immediate affective shifts triggered by specific stimuli, while \textit{Emotional Trajectory} reconstructs continuous, multi-stage emotional developments. Finally, to unravel the underlying mechanisms driving these dynamics, \textit{Emotion Cause} traces retrospectively to explain psychological motives and situational triggers, whereas \textit{Emotion Influence} projects prospectively to assess how explicit actions elicit subsequent interpersonal reactions.

\textbf{Episode-Level.} This tier evaluates long-range reasoning across full episodes or untrimmed videos, designed to assess the ability to connect temporally distant emotional cues and comprehend overarching emotional arcs. It comprises three core tasks: \textit{Emotional Intensity Comparison} requires locating temporal extrema (e.g., peak tension) or contrasting emotional magnitude across different characters, narrative moments, or emotionally relevant target subjects. \textit{Emotional Trajectory} tracks a specific character's prolonged emotional progressions and intensity dynamics spanning multiple disparate narrative scenes. Lastly, \textit{Emotional Reasoning} integrates temporally dispersed audiovisual cues across the video to either deduce discrete factual outcomes (such as reaction frequencies) or provide causal explanations for characters' deep-seated motives and long-term behavioral intentions.

\subsection{Dataset Construction and Statistics} 

To systematically construct LongEmoBench, we employ a human-in-the-loop annotation pipeline beginning with question-driven video collection. Annotators are instructed to curate long-form videos—ranging from narrative episodes to open-domain web videos—that feature rich interpersonal dynamics, pronounced emotional shifts, and sufficient narrative depth to support complex reasoning. Once a video is selected, annotators perform fine-grained audiovisual event segmentation to curate G1 question-answer pairs. To accurately capture diverse emotional dimensions, annotators employ a dual-format annotation strategy. Specifically, for tasks identifying specific emotional states (e.g., Contextual Emotion and Emotion Transition), they strictly map the emotions to the 26-class EMOTIC~\citep{Kosti_2019} taxonomy, providing a standardized and granular vocabulary for complex expressions. Conversely, for explanatory and continuous tasks (e.g., Emotion Cause and Emotion Trajectory), annotators craft natural language descriptions to articulate nuanced psychological motives and multi-stage emotional developments. Building upon these G1 annotations, annotators connect distant events across the full video to formulate G2 reasoning queries, such as tracking prolonged trajectories or comparing emotional intensity. To ensure data quality, each query is generated by one annotator and strictly validated by two independent reviewers, with any cases failing to reach a consensus strictly discarded. Further annotation details are provided in Appendix~\ref{app:annotation}.

Following this rigorous process, LongEmoBench comprises 1,354 videos totaling 69.8 hours of continuous audiovisual content, yielding 1,975 high-quality reasoning queries as demonstrated in Table~\ref{tab:benchmark_comparison}. To emphasize long-form understanding, the benchmark features extended video contexts where the G2 subset comprises full narrative episodes averaging 19.5 minutes per video. Additionally, the dataset encompasses diverse real-world content spanning sitcoms, open-domain vlogs, reality shows, live broadcasts, and various other in-the-wild formats. As illustrated in Figure~\ref{fig:statistics}, we report the statistical distributions of video durations, source domains, and 8 distinct task types.

\subsection{Evaluation Protocol}

\textbf{Label-based Evaluation.} For Contextual Emotion, Emotion Transition, and Emotion Influence, we evaluate the predicted emotional categories using the sample-averaged F1-score, following previous MER challenges~\citep{lian2025mer2025affectivecomputing}. Notably, for Emotion Transition, the F1-score is computed separately for the initial and final states and then averaged.

\textbf{Rubric-based Evaluation.} For all remaining tasks, we employ DeepSeek-V4-Flash~\citep{deepseekai2026deepseekv4} as an LLM judge to assess responses against reference answers and task-specific rubrics. Emotional Intensity Comparison and deterministic questions in Emotional Reasoning receive binary correctness scores, with semantically equivalent answers accepted. Emotional Trajectory questions receive a holistic score from 0 to 4 based on the correctness and completeness of the required emotional stages and their relationships. Questions requiring causal explanations for emotions receive a holistic score from 0 to 3, assessing explanatory correctness, sufficiency, and substantive causal errors. Scoring criteria strictly depend on the question's reasoning requirements rather than response length. To validate the reliability of this automated evaluator, we report a human-LLM correlation (Pearson's $r = 0.872$) on a sampled subset. Full rubrics are provided in Appendix~\ref{app:evaluation}.
\section{Our Method}

\begin{figure}[!t]
    \centering
    \includegraphics[width=\linewidth]{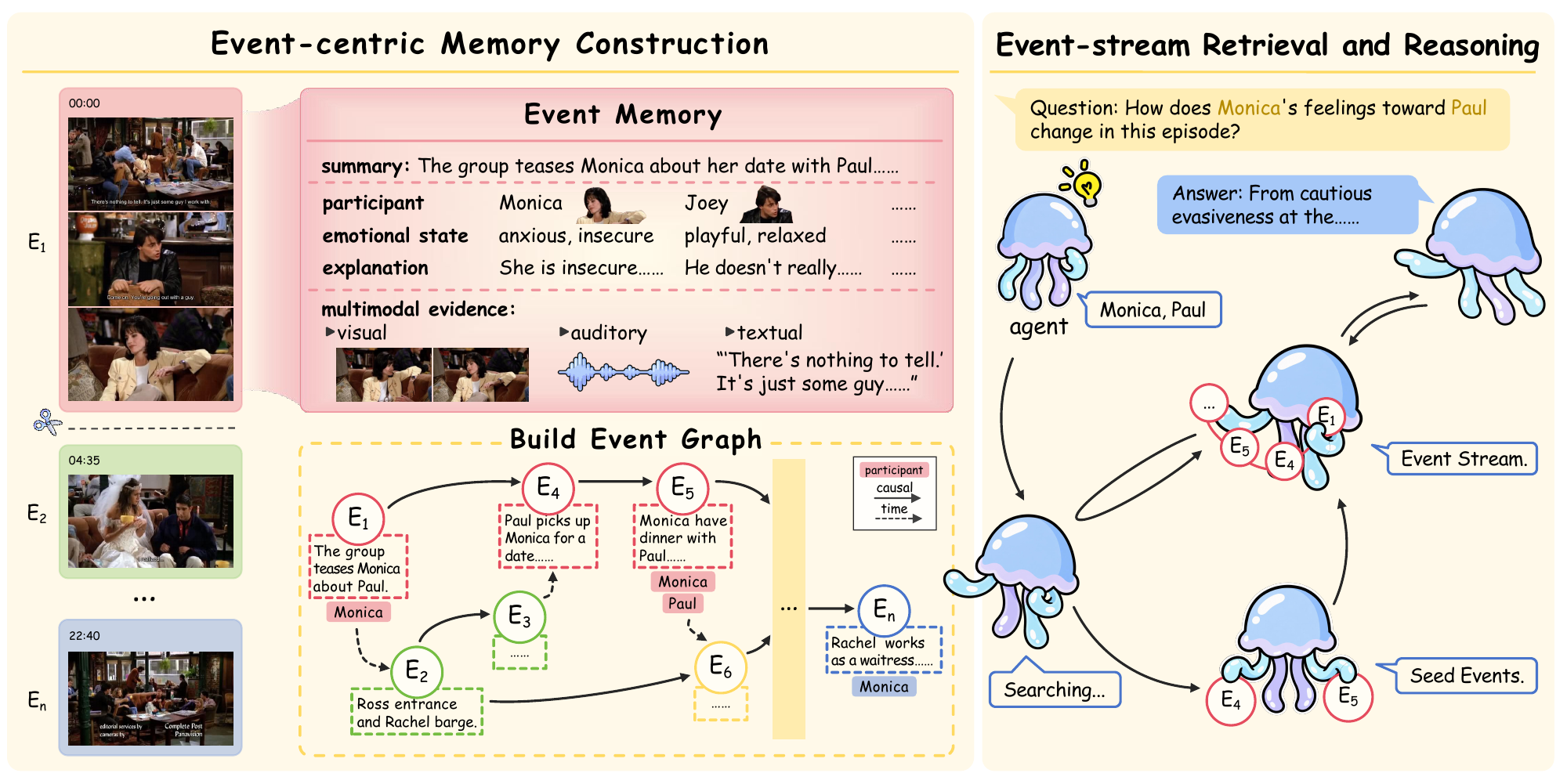}
    \caption{\textbf{Overview of our LongEmo.}}
    \label{fig:method}
\end{figure}

We introduce LongEmo, a memory-augmented framework that processes long video streams in two stages: (1) event-centric memory construction, and (2) event-stream retrieval and reasoning. As illustrated in Figure~\ref{fig:method}, we first construct event-centric memory nodes, explicitly grounding distinct emotional states in their supporting audiovisual observations and historical context (Sec~\ref{sec:4.1}). For a given query, we then retrieve a temporally ordered event stream and progressively access this evidence to execute long-range emotion reasoning and generate the answer (Sec~\ref{sec:4.2}).

\subsection{Event-Centric Memory Construction}
\label{sec:4.1}

Fundamentally, a video unfolds as a dynamic event stream within a persistent world. Human emotional states, in turn, are inherently anchored to these discrete events. Inspired by this observation, we construct our memory at the event level. To achieve this, we first instantiate discrete events as memory nodes to ground localized emotions, and subsequently construct an event graph to model long-range affective dynamics.

\textbf{Event Memory.} We formulate the global event memory as a collection of event nodes, $\mathcal{M} = \{m_1, m_2, \dots\}$, with each node defined as:
\begin{equation}
    m_i = \left(d_i, \mathcal{S}_i, \mathcal{O}_i\right)
\end{equation}
where $d_i$ is a textual summary of the event, $\mathcal{S}_i = \{(p_j, s_j, r_j)\}_{j=1}^{N_i}$ stores $N_i$ discrete emotion records, each capturing a participant $p_j$'s inferred emotional state $s_j$ and its initial causal explanation $r_j$, and $\mathcal{O}_i$ retains the supporting multimodal evidence (e.g., facial expressions, vocal prosody, and dialogue). Preserving these fine-grained raw cues is critical, as it enables the reasoning agent to subsequently reassess initial interpretations under a broader context (detailed in Sec.~\ref{sec:4.2}).

\textbf{Memory Generation and Update.} Given a long video, we process it sequentially as a series of temporal segments $\{v_t\}_{t=1}^{T}$. At step $t$, a multimodal perception module processes the current segment $v_t$ by leveraging an accumulated entity registry $\mathcal{P}_{t-1}$ to ensure consistent participant identification, alongside a recent memory subset $\mathcal{H}_{t-1} \subseteq \mathcal{M}_{t-1}$ to maintain event continuity. This process yields $K_t$ candidate event memories paired with their assignment variables $z_{t,k}$:
\begin{equation}
f_{\mathrm{perc}}\!\left(v_t, \mathcal{P}_{t-1}, \mathcal{H}_{t-1}\right)
=
\left\{(m_{t,k}, z_{t,k})\right\}_{k=1}^{K_t}
\end{equation}
Here, $z_{t,k}$ governs how each candidate is incorporated. If $z_{t,k} = \varnothing$, the candidate $m_{t,k}$ instantiates a new event node. Conversely, if $z_{t,k} = i$, it triggers an update to an ongoing event $m_i \in \mathcal{H}_{t-1}$.

During an update, the candidate's description is synthesized with the existing summary, and the new multimodal evidence is aggregated. Critically, the update mechanism for the emotion state records is designed to address complex affective dynamics through two distinct operations: \textit{sequential appending} and \textit{retrospective correction}. For genuine emotional transitions (e.g., a character shifting from ``anxiety'' to ``relief''), new states are sequentially appended to preserve the emotional progression. Conversely, when subsequent context invalidates a prior interpretation---for instance, revealing that an initial ``happy'' smile was actually a facade for ``disappointment''---the framework applies a retrospective correction to overwrite the outdated state. This context-aware revision ensures the event memory reliably reflects the underlying emotional truth rather than superficial illusions.

\textbf{Event Memory Graph Construction.} Since emotional developments are often sparsely distributed across a full episode, we construct a global event memory graph over $\mathcal{M}_T$ to capture long-range dependencies. To establish edges for each event $m_i$, the goal is to identify preceding events that provide essential historical context. However, directly querying with $m_i$ tends to retrieve analogous events. We therefore derive a targeted query $q_i$ to probe the preceding circumstances behind it. In addition, the narrative typically unfolds through events centered on recurring characters, making shared participants a strong indicator of cross-event relevance. To this end, we retrieve historical events by jointly considering query relevance and character continuity. Formally, this retrieval process is formulated as:
\begin{equation}
s(m_h, m_i)
=
\operatorname{rel}(q_i, m_h)
+
\lambda_p
\frac{\lvert \mathcal{P}_i \cap \mathcal{P}_h \rvert}
     {\lvert \mathcal{P}_i \cup \mathcal{P}_h \rvert},
\quad h < i.
\end{equation}
where $\operatorname{rel}(q_i, m_h)$ evaluates the relevance between query $q_i$ and candidate event $m_h$ via a hybrid retrieval scheme combining dense semantic embeddings with BM25 lexical matching, $\mathcal{P}_i$ and $\mathcal{P}_h$ denote the character sets of $m_i$ and $m_h$, respectively, and $\lambda_p$ is a trade-off coefficient. Preceding events are accordingly ranked by $s(m_h, m_i)$ to form a candidate pool $\mathcal{C}_i$. To eliminate spurious dependencies, a relation verifier $f_{\mathrm{ver}}$ subsequently inspects each candidate in $\mathcal{C}_i$ to establish valid dependencies, yielding the final event memory graph:
\begin{equation}
\mathcal{G} = (\mathcal{M}_T, \mathcal{R}), \qquad
\mathcal{R} = \bigcup_{i=1}^{T}
\left\{
e_{h \rightarrow i}
\;\middle|\;
m_h \in \mathcal{C}_i,\;
f_{\mathrm{ver}}(m_h, m_i) = 1
\right\}.
\end{equation}

\subsection{Event-Stream Retrieval and Reasoning}
\label{sec:4.2} 

Directly injecting the entire event memory graph $\mathcal{G}$ into the reasoning model introduces prohibitive computational redundancy and context noise. To address this, we first retrieve a focused set of seed events and expand them along the graph edges into a chronologically ordered event stream, which serves as a structured scaffold for progressive reasoning.

\textbf{Seed Event Retrieval.} Given a question $q$, we identify high-confidence entry events by applying the same hybrid retrieval and character continuity scheme defined in Sec.~\ref{sec:4.1}. Specifically, we extract queried characters $\mathcal{P}_q$ from $q$ and rank all events in $\mathcal{M}_T$ by jointly scoring text relevance and Jaccard character similarity with trade-off coefficient $\lambda_q$. The top-$K$ candidates are retained to form the initial seed set, establishing temporal anchors along the video timeline.

\textbf{Relational Event-Stream Expansion.} While seed events provide reliable entry points, they remain isolated moments that cannot capture emotional dynamics on their own. We therefore expand the seed set by retrieving connected events along relational edges in the event memory graph. The expanded event set is then sorted chronologically into an ordered event stream. To respect the reasoning model's context budget, we exclude raw multimodal signals and provide only each event's timestamp, textual summary, participating characters, and relational links as a lightweight overview.

\textbf{Progressive Multimodal Reasoning.} With the lightweight event stream providing global context, the model conducts targeted multimodal inspection rather than indiscriminately processing the entire video. It first scans chronological text summaries to trace character interactions and locate critical turning points relevant to the query. For these pivotal moments, the model selectively unpacks their fine-grained emotional annotations alongside the corresponding audiovisual clips. By integrating local multimodal cues, such as vocal tone and facial expressions, with the global narrative trajectory, the model resolves emotional ambiguities and derives a grounded, context-aware answer.

\begin{table*}[t]
\centering
\setlength{\belowcaptionskip}{\baselineskip}

\caption{G1 results (\%) on LongEmoBench. Rubric scores for specific tasks (e.g., 0--4 for Trajectory and 0--3 for Cause) are converted to percentages for uniform comparison. V, A, and T denote visual, audio, and textual inputs, respectively. Overall denotes the weighted average.}
\label{tab:scene_results}

\setlength{\tabcolsep}{2.6pt}
\renewcommand{\arraystretch}{1.0}

\begin{tabularx}{\textwidth}{
    @{}X
    c c c@{\hspace{10pt}}
    *{6}{c}@{}
}
\toprule

\textbf{Method}
& \textbf{V}
& \textbf{A}
& \textbf{T}
& \textbf{Contextual}
& \textbf{Transition}
& \textbf{Trajectory}
& \textbf{Cause}
& \textbf{Influence}
& \textbf{Overall} \\

\midrule

\rowcolor{LEGroupBg}[0pt][0pt]
\multicolumn{10}{@{}l@{}}{
    \hspace{4pt}
    \textcolor{LEAccent}{\strut\bfseries Affective MLLMs}
} \\

R1-Omni
& \LEyes & \LEyes & \LEyes
& 0.05 & 0.00 & 12.05 & 2.20 & 1.38 & 5.38 \\

HumanOmni
& \LEyes & \LEyes & \LEyes
& 1.40 & 0.00 & 12.23 & 5.82 & 13.01 & 9.12 \\

AffectGPT
& \LEyes & \LEyes & \LEyes
& 13.20 & 15.65 & 16.52 & 12.71 & 15.97 & 15.32 \\

Emotion-LLaMA
& \LEyes & \LEyes & \LEyes
& 9.51 & 12.00 & 13.59 & 9.75 & 24.03 & 15.03 \\

\addlinespace[3pt]

\rowcolor{LEGroupBg}[0pt][0pt]
\multicolumn{10}{@{}l@{}}{
    \hspace{4pt}
    \textcolor{LEAccent}{\strut\bfseries General-Purpose MLLMs}
} \\

Qwen2-Audio-7B
& \LEno & \LEyes & \LEyes
& 5.57 & 1.68 & 22.92 & 27.67 & 8.81 & 16.14 \\

Qwen3-VL-8B
& \LEyes & \LEno & \LEyes
& 31.52 & 29.57 & 45.43 & 60.06 & 33.97 & 41.67 \\

Qwen3-Omni-30B
& \LEyes & \LEyes & \LEyes
& 36.38 & 38.61 & 47.83 & 68.24 & 40.43 & 46.82 \\

InternVL3.5-38B
& \LEyes & \LEno & \LEyes
& 36.98 & 34.26 & 37.18 & 55.97 & 40.52 & 40.58 \\

Gemini-3-Flash
& \LEyes & \LEyes & \LEyes
& \textbf{59.07}
& \textbf{57.89}
& \textbf{63.41}
& \textbf{87.58}
& \textbf{58.36}
& \textbf{64.75} \\

\addlinespace[3pt]

\rowcolor{LEGroupBg}[0pt][0pt]
\multicolumn{10}{@{}l@{}}{
    \hspace{4pt}
    \textcolor{LEAccent}{\strut\bfseries Long-Video Architectures}
} \\

LongVU
& \LEyes & \LEno & \LEyes
& 8.06 & 19.90 & 24.86 & 34.91 & 19.96 & 22.61 \\

Flash-VStream-7B
& \LEyes & \LEno & \LEyes
& 5.28 & 0.00 & 13.59 & 24.84 & 7.09 & 11.47 \\

VideoChat-Flash-7B
& \LEyes & \LEno & \LEyes
& 31.86 & 24.12 & 32.20 & 38.52 & 34.43 & 33.01 \\

VideoChat3-4B
& \LEyes & \LEno & \LEyes
& 35.43 & 35.92 & 38.77 & 49.21 & 36.75 & 39.17 \\

\bottomrule
\end{tabularx}
\end{table*}

\section{Experiments}

\textbf{Experimental Setup.} 
For G1 tasks, we assess general-purpose MLLMs (Qwen3-VL-8B, Qwen2-Audio-7B, Qwen3-Omni-30B, InternVL3.5-38B, Gemini-3-Flash), affective MLLMs (R1-Omni~\citep{zhao2025r1omni}, HumanOmni~\citep{zhao2025humanomni}, AffectGPT~\citep{lian2025affectgpt}, Emotion-LLaMA~\citep{cheng2024emotionllama}), and long-video architectures (LongVU~\citep{shen2024longvu}, Flash-VStream~\citep{zhang2025flashvstream}, VideoChat-Flash~\citep{li2025videochatflash}, VideoChat3~\citep{li2026videochat3}). For G2 tasks, we evaluate the long-video architectures alongside agentic systems (VideoHV~\citep{wang2026videohv}, LongVideoAgent~\citep{liu2025longvideoagent}, M3-Agent~\citep{long2025m3agent}, WorldMM~\citep{yeo2026worldmm}). All implementation details are provided in the Appendix~\ref{app:implementation}.

\begin{table}[t]
\centering
\begin{minipage}[c]{0.56\textwidth}
\caption{G2 results (\%) on LongEmoBench. Task-specific rubric scores are converted to percentages for uniform comparison. EIC: Emotional Intensity Comparison; ET: Emotional Trajectory; ER: Emotional Reasoning. Overall denotes the weighted average.}
\label{tab:episode_results}
\vspace{5pt}
\setlength{\tabcolsep}{4pt}
\renewcommand{\arraystretch}{1.0}
\begin{tabularx}{\linewidth}{@{}>{\raggedright\arraybackslash}Xcccc@{}}
\toprule
\textbf{Method} & \textbf{EIC} & \textbf{ET} & \textbf{ER} & \textbf{Overall} \\
\midrule
\rowcolor{LEGroupBg}[0pt][0pt]
\multicolumn{5}{@{}l@{}}{\strut\color{LEAccent}\bfseries Long-Video Architectures} \\
LongVU & 1.55 & 23.94 & 20.75 & 15.42 \\
Flash-VStream-7B & 4.64 & 9.89 & 8.48 & 7.74 \\
VideoChat-Flash-7B & 1.55 & 30.64 & 21.46 & 18.40 \\
VideoChat3-4B & 10.31 & 30.21 & 26.46 & 22.42 \\
\addlinespace[3pt]
\rowcolor{LEGroupBg}[0pt][0pt]
\multicolumn{5}{@{}l@{}}{\strut\color{LEAccent}\bfseries Agent Methods} \\
VideoHV & 2.06 & 34.89 & 38.50 & 24.31 \\
LongVideoAgent & 7.73 & 21.91 & 11.89 & 14.67 \\
M3-Agent & 8.76 & 31.06 & 28.94 & 22.82 \\
WorldMM & 29.38 & 51.49 & 58.66 & 45.46 \\
\midrule
\textbf{LongEmo (Ours)} & \textbf{49.13} & \textbf{60.60} & \textbf{75.48} & \textbf{60.24} \\
\bottomrule
\end{tabularx}
\end{minipage}\hfill
\begin{minipage}[c]{0.42\textwidth}
\centering
\begin{tikzpicture}
\begin{axis}[
  scale only axis,
  width=\dimexpr\linewidth-20pt\relax,
  height=137pt,
  ybar=1pt,
  bar width=7.5pt,
  ymin=0, ymax=85,
  ytick={0,20,40,60,80},
  yticklabels={0,20,40,60,80},
  symbolic x coords={EIC,ET,ER,Overall},
  xtick=data,
  enlarge x limits=0.19,
  axis x line*=bottom,
  axis y line*=left,
  y axis line style={draw=none},
  axis line style={black!35,line width=0.4pt},
  tick style={draw=none},
  major tick length=0pt,
  ymajorgrids=true,
  grid style={black!10,line width=0.3pt},
  xticklabel style={font=\small,text=black!90,inner sep=3pt},
  yticklabel style={font=\small,text=black!65,inner sep=2pt},
  legend style={
    at={(0,1.055)},anchor=south west,
    draw=none,fill=none,font=\small,
    legend columns=1,row sep=1.5pt,
    inner sep=0pt,
    /tikz/every even column/.append style={column sep=3pt},
  },
  legend cell align=left,
  legend image code/.code={\path[#1] (0cm,-0.075cm) rectangle (0.18cm,0.075cm);},
  clip=true,
]
\addplot[fill=LEAblMemory,draw=LEAblMemory!80!black,line width=0.25pt]
  coordinates {(EIC,39.66) (ET,53.67) (ER,65.81) (Overall,51.64)};
\addlegendentry{w/o Event Memory}
\addplot[fill=LEAblRetrieval,draw=LEAblRetrieval!80!black,line width=0.25pt]
  coordinates {(EIC,51.15) (ET,54.41) (ER,72.68) (Overall,57.62)};
\addlegendentry{w/o Event-Stream Retrieval}
\addplot[fill=LEAblLongEmo,draw=LEAblLongEmo!80!black,line width=0.25pt]
  coordinates {(EIC,49.13) (ET,60.60) (ER,75.48) (Overall,60.24)};
\addlegendentry{LongEmo}
\end{axis}
\end{tikzpicture}

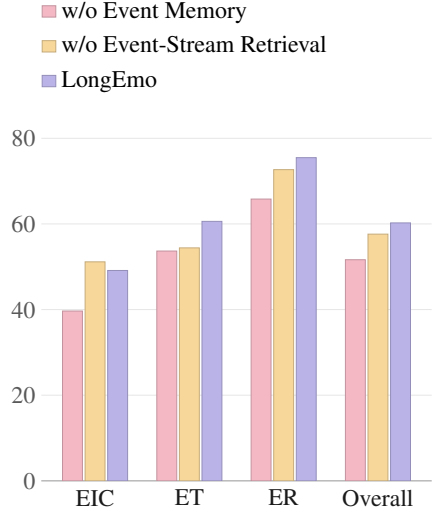
\captionof{figure}{Ablation results of LongEmo.}
\label{fig:longemo_ablation}
\end{minipage}
\end{table}

\subsection{Main Results}

Tables~\ref{tab:scene_results} and~\ref{tab:episode_results} report the results on the G1 and G2 tasks of LongEmoBench, respectively.

\textbf{Results on G1.} The G1 evaluation yields three critical observations: 

(1) Affective MLLMs exhibit surprisingly limited capabilities, with AffectGPT and Emotion-LLaMA achieving only 15.32\% and 15.03\%, significantly trailing general-purpose MLLMs like Qwen3-Omni-30B. This indicates that specialized fine-tuning within the affective computing domain, which primarily targets static or short-clip recognition, inadvertently compromises the broad contextual reasoning capabilities required for complex emotion reasoning in dynamic scenes. 

(2) Among the G1 tasks, most models perform best on Emotion Cause, with Gemini-3-Flash and Qwen3-Omni-30B scoring 87.58\% and 68.24\%, respectively. In contrast, tasks requiring continuous emotional reasoning across the entire video, such as Contextual Emotion and Emotion Transition, prove significantly more challenging, with the top-performing Gemini-3-Flash reaching only 59.07\% and 57.89\%, respectively. We attribute this phenomenon to the fact that current models excel at factual reasoning, making the deduction of emotion triggers a straightforward task for them. However, they fundamentally lack the capability to capture holistic emotional states and track fluid dynamics across the complete emotion lifecycle. 

(3) Long-video models trained on general scenarios struggle significantly with these emotion tasks. Even the leading VideoChat3-4B achieves only 39.17\%, consistently trailing general-purpose MLLMs. This implies that the temporal compression and frame-subsampling strategies inherent to long-video architectures inadvertently discard the fine-grained multimodal cues essential for subtle emotion~perception, as these mechanisms are fundamentally optimized to retain high-level semantic actions rather than fleeting affective details.

\textbf{Results on G2.} These results highlights two key findings regarding long-video emotion reasoning: 

(1) Long-video architectures exhibit even more severe deficiencies on extended videos, with their overall scores dropping further to around 20\%. The agentic video understanding methods VideoHV and LongVideoAgent, which autonomously navigate and sample video segments, suffer from these exact same limitations. This indicates that these paradigms fundamentally fail to preserve the continuous, fine-grained temporal cues required for robust emotion reasoning over long contexts.

(2) Memory-augmented strategies emerge as a promising direction to address these limitations. Notably, WorldMM achieves a strong score of 45.46\%. We attribute this success to its capacity for cross-temporal episodic memory construction, a critical design philosophy it shares with our LongEmo. This need is further highlighted by M3-Agent scoring only 22.82\% due to its restrictive fixed-clip memory span. Building upon this, LongEmo utilizes an event-centric memory graph to anchor subtle emotional shifts to explicit narrative events, enabling precise cross-context retrieval and driving a 14.78-point improvement over the strongest baseline.

\subsection{Ablation Study}

Figure~\ref{fig:longemo_ablation} details the evaluation of two architectural variants. \textit{w/o Event Memory} bypasses event-level memory construction, instead relying on fixed-window descriptions that are retrieved directly for each question. This degrades the Overall score from 60.24\% to 51.64\% across all tasks, confirming that preserving emotional states within coherent events is superior to strict temporal boundaries. \textit{w/o Event-Stream Retrieval} retrieves events independently without expanding graph relations. Consequently, the Overall score drops to 57.62\%, with ET and ER decreasing to 54.41\% and 72.68\%, respectively. This demonstrates that isolated events lack sufficient context for tracking emotional evolution. By expanding seed events relationally, LongEmo effectively bridges historical context and local evidence, confirming the need for contextual continuity in long-range emotion reasoning.

\begin{figure}[!t]
\centering
\begin{minipage}[t]{0.64\textwidth}
  \vspace{0pt}\centering
  \includegraphics[width=\linewidth]{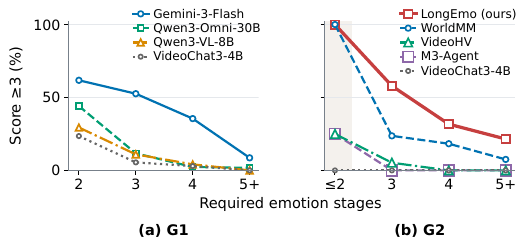}
  \caption{Impact of emotional dynamics complexity.}
  \label{fig:trajectory_stages}
\end{minipage}\hfill
\begin{minipage}[t]{0.34\textwidth}
  \vspace{0pt}\centering
  \includegraphics[width=\linewidth]{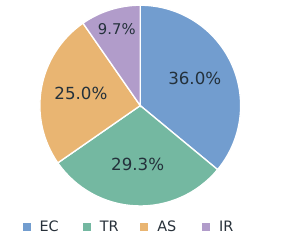}
  \caption{Failure diagnosis.}
  \label{fig:failure_attribution}
\end{minipage}
\end{figure}

\subsection{More Analysis}
\label{sec:more_analysis}

\textbf{Impact of emotional dynamics complexity.}
We measure the complexity of emotional dynamics using the number of trajectory stages. As Figure~\ref{fig:trajectory_stages} shows, the ability of all methods to capture complete trajectories drops across both granularities as the stage count increases. This trend highlights a fundamental limitation of existing methods in tracking continuous emotional dynamics. Specifically, critical intermediate emotional states are easily diluted or overwritten within the expanding multimodal context, breaking the causal chain of the trajectory.

\textbf{Failure diagnosis.}
We classify LongEmo's G2 failure cases into four types of deficiencies (Figure~\ref{fig:failure_attribution}): insufficient evidence acquisition and coverage (EC), flawed task constraints and reasoning (TR), inadequate answer synthesis and compression (AS), and incorrect information interpretation and representation (IR). Most errors stem from EC and AS, showing the model struggles to retrieve complete evidence and compress complex emotional stages. This highlights the critical need to enhance memory comprehensiveness and emotional state fidelity.

\section{Conclusion}

\textbf{Limitations and Social Impact.} LongEmoBench may underrepresent subtle and culturally diverse emotional behavior, while emotion annotations can admit multiple reasonable interpretations. This work may inform research on context-sensitive video systems. Emotional predictions should not be treated as objective measurements of internal states or grounds for high-stakes decisions.

\textbf{Summary and Outlook.} We introduced LongEmoBench and LongEmo for emotion understanding across scenes and episodes. Our findings support integrating local multimodal evidence with long-range context through structured event memories and relational retrieval. Future work should expand data and task diversity, develop uncertainty-aware memory and retrieval, and incorporate diverse human judgments to improve the reliability and generalization of long-range emotional reasoning.

\subsection*{AI USE STATEMENT}
Generative AI tools were used solely to polish the manuscript’s language, including improvements to grammar, clarity, and readability. The authors take full responsibility for the scientific content and final wording of the manuscript.
\subsection*{ETHICS STATEMENT}
This study does not involve human participants or sensitive personal data. We do not identify specific ethical concerns arising from this work. Downstream applications should be evaluated in their intended contexts, with consideration for fairness, privacy, and potential misuse.
\subsection*{REPRODUCIBILITY STATEMENT}
The main text and appendix describe the proposed method, datasets, experimental settings, and evaluation procedures. Additional implementation details and hyperparameter settings are provided to support reproduction of the reported results.

\bibliography{iclr2027_conference}

\begin{thebibliography}{41}
\providecommand{\natexlab}[1]{#1}
\providecommand{\url}[1]{\texttt{#1}}
\expandafter\ifx\csname urlstyle\endcsname\relax
  \providecommand{\doi}[1]{doi: #1}\else
  \providecommand{\doi}{doi: \begingroup \urlstyle{rm}\Url}\fi

\bibitem[Cao et~al.(2026)Cao, Zhang, Yu, Zhang, Cao, Lu, Lin, Han, and Sun]{cao2026omnifocus}
Shijie Cao, Qingyu Zhang, Boxi Yu, Yuzhong Zhang, Boxi Cao, Yaojie Lu, Hongyu Lin, Xianpei Han, and Le~Sun.
\newblock Omnifocus: Query-guided modality-balanced token compression for omni-modal large language models, 2026.
\newblock URL \url{https://arxiv.org/abs/2607.03050}.

\bibitem[Chen et~al.(2026)Chen, Tan, Yu, Wang, Cheng, Guan, Jiang, Li, Zhu, and Song]{chen2026avoc}
Yijing Chen, Wenhui Tan, Xiaoyi Yu, Yuyue Wang, Xin Cheng, Kaisi Guan, Hao Jiang, Xiangyang Li, Guojie Zhu, and Ruihua Song.
\newblock Avoc: Enhancing hour-level audio-video understanding in omni-modal llms via retrieval-inspired token compression, 2026.
\newblock URL \url{https://arxiv.org/abs/2606.24286}.

\bibitem[Chen \& Whitney(2019)Chen and Whitney]{chen2019tracking}
Zhimin Chen and David Whitney.
\newblock Tracking the affective state of unseen persons.
\newblock \emph{Proceedings of the National Academy of Sciences}, 116\penalty0 (15):\penalty0 7559--7564, 2019.

\bibitem[Cheng et~al.(2024)Cheng, Cheng, He, Sun, Wang, Lin, Lian, Peng, and Hauptmann]{cheng2024emotionllama}
Zebang Cheng, Zhi-Qi Cheng, Jun-Yan He, Jingdong Sun, Kai Wang, Yuxiang Lin, Zheng Lian, Xiaojiang Peng, and Alexander Hauptmann.
\newblock Emotion-llama: Multimodal emotion recognition and reasoning with instruction tuning, 2024.
\newblock URL \url{https://arxiv.org/abs/2406.11161}.

\bibitem[DeepSeek-AI(2026)]{deepseekai2026deepseekv4}
DeepSeek-AI.
\newblock Deepseek-v4: Towards highly efficient million-token context intelligence, 2026.
\newblock URL \url{https://arxiv.org/abs/2606.19348}.

\bibitem[Fu et~al.(2025)Fu, Dai, Luo, Li, Ren, Zhang, Wang, Zhou, Shen, Zhang, Chen, Li, Lin, Zhao, Li, Xu, Zheng, Chen, Shan, He, and Sun]{fu2025videomme}
Chaoyou Fu, Yuhan Dai, Yongdong Luo, Lei Li, Shuhuai Ren, Renrui Zhang, Zihan Wang, Chenyu Zhou, Yunhang Shen, Mengdan Zhang, Peixian Chen, Yanwei Li, Shaohui Lin, Sirui Zhao, Ke~Li, Tong Xu, Xiawu Zheng, Enhong Chen, Caifeng Shan, Ran He, and Xing Sun.
\newblock Video-mme: The first-ever comprehensive evaluation benchmark of multi-modal llms in video analysis, 2025.
\newblock URL \url{https://arxiv.org/abs/2405.21075}.

\bibitem[Gross \& Levenson(1993)Gross and Levenson]{gross1993emotional}
James~J Gross and Robert~W Levenson.
\newblock Emotional suppression: physiology, self-report, and expressive behavior.
\newblock \emph{Journal of personality and social psychology}, 64\penalty0 (6):\penalty0 970, 1993.

\bibitem[Houlihan et~al.(2023)Houlihan, Kleiman-Weiner, Hewitt, Tenenbaum, and Saxe]{houlihan2023emotion}
Sean~Dae Houlihan, Max Kleiman-Weiner, Luke~B Hewitt, Joshua~B Tenenbaum, and Rebecca Saxe.
\newblock Emotion prediction as computation over a generative theory of mind.
\newblock \emph{Philosophical transactions. Series A, Mathematical, physical, and engineering sciences}, 381\penalty0 (2251):\penalty0 20220047, 2023.

\bibitem[Hu et~al.(2026{\natexlab{a}})Hu, Weng, Cheng, Wang, Luo, Schuller, Lian, and Cui]{hu2026emotrans}
He~Hu, Tengjin Weng, Zebang Cheng, Yu~Wang, Jiachen Luo, Björn Schuller, Zheng Lian, and Laizhong Cui.
\newblock Emotrans: A benchmark for understanding, reasoning, and predicting emotion transitions in multimodal llms, 2026{\natexlab{a}}.
\newblock URL \url{https://arxiv.org/abs/2604.23348}.

\bibitem[Hu et~al.(2026{\natexlab{b}})Hu, You, Xu, Wang, Yu, Ma, Cheng, Lian, Zhou, and Cui]{hu2026emobenchm}
He~Hu, Lianzhong You, Hongbo Xu, Qianning Wang, Fei~Richard Yu, Fei Ma, Zebang Cheng, Zheng Lian, Yucheng Zhou, and Laizhong Cui.
\newblock Emobench-m: Benchmarking emotional intelligence for multimodal large language models, 2026{\natexlab{b}}.
\newblock URL \url{https://arxiv.org/abs/2502.04424}.

\bibitem[Hu et~al.(2025)Hu, Shi, Dai, Li, Song, and Wang]{hu2025mtmeur}
Jinpeng Hu, Hongchang Shi, Chongyuan Dai, Zhuo Li, Peipei Song, and Meng Wang.
\newblock Beyond emotion recognition: A multi-turn multimodal emotion understanding and reasoning benchmark, 2025.
\newblock URL \url{https://arxiv.org/abs/2508.16859}.

\bibitem[Kosti et~al.(2019)Kosti, Alvarez, Recasens, and Lapedriza]{Kosti_2019}
Ronak Kosti, Jose Alvarez, Adria Recasens, and Agata Lapedriza.
\newblock Context based emotion recognition using emotic dataset.
\newblock \emph{IEEE Transactions on Pattern Analysis and Machine Intelligence}, pp.\  1–1, 2019.
\newblock ISSN 1939-3539.
\newblock \doi{10.1109/tpami.2019.2916866}.
\newblock URL \url{http://dx.doi.org/10.1109/TPAMI.2019.2916866}.

\bibitem[Krippendorff(2011)]{krippendorff2011alpha}
Klaus Krippendorff.
\newblock Computing {Krippendorff}'s alpha-reliability.
\newblock Technical report, Annenberg School for Communication, University of Pennsylvania, 2011.
\newblock URL \url{https://www.asc.upenn.edu/sites/default/files/2021-03/Computing%20Krippendorff%27s%20Alpha-Reliability.pdf}.

\bibitem[Li et~al.(2025)Li, Wang, Yu, Zeng, Zhu, Huang, Gao, Li, He, Wang, Qiao, Wang, and Wang]{li2025videochatflash}
Xinhao Li, Yi~Wang, Jiashuo Yu, Xiangyu Zeng, Yuhan Zhu, Haian Huang, Jianfei Gao, Kunchang Li, Yinan He, Chenting Wang, Yu~Qiao, Yali Wang, and Limin Wang.
\newblock Videochat-flash: Hierarchical compression for long-context video modeling, 2025.
\newblock URL \url{https://arxiv.org/abs/2501.00574}.

\bibitem[Li et~al.(2026)Li, Zhu, Zeng, Dong, Wu, Zhang, Yang, Ma, Zhang, Shi, Chen, Chen, Huang, Zhang, Ouyang, Sui, Yan, Xu, Wang, He, Zhang, Wang, Qiao, Wang, Liu, Chen, and Wang]{li2026videochat3}
Xinhao Li, Yuhan Zhu, Xiangyu Zeng, Yuhao Dong, Haoning Wu, Zhiqiu Zhang, Yuandong Yang, Changlian Ma, Qingyu Zhang, Yansong Shi, Xinyu Chen, Haoran Chen, Zizheng Huang, Jun Zhang, Kun Ouyang, Lin Sui, Ziang Yan, Yicheng Xu, Chenting Wang, Yinan He, Hongjie Zhang, Yi~Wang, Yu~Qiao, Yali Wang, Ziwei Liu, Kai Chen, and Limin Wang.
\newblock Videochat3: Fully open video mllm for efficient and generalist video understanding, 2026.
\newblock URL \url{https://arxiv.org/abs/2607.14935}.

\bibitem[Lian et~al.(2024{\natexlab{a}})Lian, Sun, Sun, Wen, Zhang, Chen, Gu, Zhao, Ma, Chen, Yi, Liu, Xu, Liu, Cambria, Zhao, Schuller, and Tao]{lian2024mer2024}
Zheng Lian, Haiyang Sun, Licai Sun, Zhuofan Wen, Siyuan Zhang, Shun Chen, Hao Gu, Jinming Zhao, Ziyang Ma, Xie Chen, Jiangyan Yi, Rui Liu, Kele Xu, Bin Liu, Erik Cambria, Guoying Zhao, Björn~W. Schuller, and Jianhua Tao.
\newblock Mer 2024: Semi-supervised learning, noise robustness, and open-vocabulary multimodal emotion recognition, 2024{\natexlab{a}}.
\newblock URL \url{https://arxiv.org/abs/2404.17113}.

\bibitem[Lian et~al.(2024{\natexlab{b}})Lian, Sun, Ren, Gu, Sun, Chen, Liu, and Tao]{lian2024merbench}
Zheng Lian, Licai Sun, Yong Ren, Hao Gu, Haiyang Sun, Lan Chen, Bin Liu, and Jianhua Tao.
\newblock Merbench: A unified evaluation benchmark for multimodal emotion recognition, 2024{\natexlab{b}}.
\newblock URL \url{https://arxiv.org/abs/2401.03429}.

\bibitem[Lian et~al.(2025{\natexlab{a}})Lian, Chen, Chen, Sun, Sun, Ren, Cheng, Liu, Liu, Peng, Yi, and Tao]{lian2025affectgpt}
Zheng Lian, Haoyu Chen, Lan Chen, Haiyang Sun, Licai Sun, Yong Ren, Zebang Cheng, Bin Liu, Rui Liu, Xiaojiang Peng, Jiangyan Yi, and Jianhua Tao.
\newblock Affectgpt: A new dataset, model, and benchmark for emotion understanding with multimodal large language models, 2025{\natexlab{a}}.
\newblock URL \url{https://arxiv.org/abs/2501.16566}.

\bibitem[Lian et~al.(2025{\natexlab{b}})Lian, Liu, Xu, Liu, Liu, Zhang, Liu, Li, Cheng, Zuo, Ma, Peng, Chen, Li, Cambria, Zhao, Schuller, and Tao]{lian2025mer2025}
Zheng Lian, Rui Liu, Kele Xu, Bin Liu, Xuefei Liu, Yazhou Zhang, Xin Liu, Yong Li, Zebang Cheng, Haolin Zuo, Ziyang Ma, Xiaojiang Peng, Xie Chen, Ya~Li, Erik Cambria, Guoying Zhao, Björn~W. Schuller, and Jianhua Tao.
\newblock Mer 2025: When affective computing meets large language models, 2025{\natexlab{b}}.
\newblock URL \url{https://arxiv.org/abs/2504.19423}.

\bibitem[Lian et~al.(2025{\natexlab{c}})Lian, Liu, Xu, Liu, Liu, Zhang, Liu, Li, Cheng, Zuo, Ma, Peng, Chen, Li, Cambria, Zhao, Schuller, and Tao]{lian2025mer2025affectivecomputing}
Zheng Lian, Rui Liu, Kele Xu, Bin Liu, Xuefei Liu, Yazhou Zhang, Xin Liu, Yong Li, Zebang Cheng, Haolin Zuo, Ziyang Ma, Xiaojiang Peng, Xie Chen, Ya~Li, Erik Cambria, Guoying Zhao, Björn~W. Schuller, and Jianhua Tao.
\newblock Mer 2025: When affective computing meets large language models, 2025{\natexlab{c}}.
\newblock URL \url{https://arxiv.org/abs/2504.19423}.

\bibitem[Lian et~al.(2025{\natexlab{d}})Lian, Sun, Sun, Chen, Chen, Gu, Wen, Chen, Zhang, Yao, Liu, Liu, Liang, Li, Yi, and Tao]{lian2025ovmer}
Zheng Lian, Haiyang Sun, Licai Sun, Haoyu Chen, Lan Chen, Hao Gu, Zhuofan Wen, Shun Chen, Siyuan Zhang, Hailiang Yao, Bin Liu, Rui Liu, Shan Liang, Ya~Li, Jiangyan Yi, and Jianhua Tao.
\newblock Ov-mer: Towards open-vocabulary multimodal emotion recognition, 2025{\natexlab{d}}.
\newblock URL \url{https://arxiv.org/abs/2410.01495}.

\bibitem[Lin et~al.(2026)Lin, Tong, Wu, Zhang, Liu, Jin, and Shen]{lin2026speakwatching}
Junyan Lin, Junlong Tong, Hao Wu, Jialiang Zhang, Jinming Liu, Xin Jin, and Xiaoyu Shen.
\newblock Speak while watching: Unleashing true real-time video understanding capability of multimodal large language models, 2026.
\newblock URL \url{https://arxiv.org/abs/2601.06843}.

\bibitem[Liu et~al.(2025)Liu, Liu, Tang, Ma, Pi, Zhang, and Chen]{liu2025longvideoagent}
Runtao Liu, Ziyi Liu, Jiaqi Tang, Yue Ma, Renjie Pi, Jipeng Zhang, and Qifeng Chen.
\newblock Longvideoagent: Multi-agent reasoning with long videos, 2025.
\newblock URL \url{https://arxiv.org/abs/2512.20618}.

\bibitem[Long et~al.(2025)Long, He, Ye, Pan, Lin, Li, Zhao, and Li]{long2025m3agent}
Lin Long, Yichen He, Wentao Ye, Yiyuan Pan, Yuan Lin, Hang Li, Junbo Zhao, and Wei Li.
\newblock Seeing, listening, remembering, and reasoning: A multimodal agent with long-term memory, 2025.
\newblock URL \url{https://arxiv.org/abs/2508.09736}.

\bibitem[Nie et~al.(2026)Nie, Wei, Feng, Fu, and Shan]{nie2026lightomni}
Chang Nie, Jiaju Wei, Junlan Feng, Chaoyou Fu, and Caifeng Shan.
\newblock Light-omni: Reflex over reasoning in agentic video understanding with long-term memory, 2026.
\newblock URL \url{https://arxiv.org/abs/2607.05511}.

\bibitem[Ong et~al.(2015)Ong, Zaki, and Goodman]{ong2015affective}
Desmond~C Ong, Jamil Zaki, and Noah~D Goodman.
\newblock Affective cognition: Exploring lay theories of emotion.
\newblock \emph{Cognition}, 143:\penalty0 141--162, 2015.

\bibitem[Sasu et~al.(2025)Sasu, Wu, Gong, Chen, Shi, Ai, Hirschberg, and Schluter]{sasu-etal-2025-akan}
David Sasu, Zehui Wu, Ziwei Gong, Run Chen, Pengyuan Shi, Lin Ai, Julia Hirschberg, and Natalie Schluter.
\newblock {A}kan cinematic emotions ({ACE}): A multimodal multi-party dataset for emotion recognition in movie dialogues.
\newblock In Wanxiang Che, Joyce Nabende, Ekaterina Shutova, and Mohammad~Taher Pilehvar (eds.), \emph{Findings of the Association for Computational Linguistics: ACL 2025}, pp.\  9820--9831, Vienna, Austria, July 2025. Association for Computational Linguistics.
\newblock ISBN 979-8-89176-256-5.
\newblock \doi{10.18653/v1/2025.findings-acl.510}.
\newblock URL \url{https://aclanthology.org/2025.findings-acl.510/}.

\bibitem[Scherer(2009)]{scherer2009emotions}
Klaus~R Scherer.
\newblock Emotions are emergent processes: they require a dynamic computational architecture.
\newblock \emph{Philosophical Transactions of the Royal Society B: Biological Sciences}, 364\penalty0 (1535):\penalty0 3459, 2009.

\bibitem[Shen et~al.(2024)Shen, Xiong, Zhao, Wu, Chen, Zhu, Liu, Xiao, Varadarajan, Bordes, Liu, Xu, Kim, Soran, Krishnamoorthi, Elhoseiny, and Chandra]{shen2024longvu}
Xiaoqian Shen, Yunyang Xiong, Changsheng Zhao, Lemeng Wu, Jun Chen, Chenchen Zhu, Zechun Liu, Fanyi Xiao, Balakrishnan Varadarajan, Florian Bordes, Zhuang Liu, Hu~Xu, Hyunwoo~J. Kim, Bilge Soran, Raghuraman Krishnamoorthi, Mohamed Elhoseiny, and Vikas Chandra.
\newblock Longvu: Spatiotemporal adaptive compression for long video-language understanding, 2024.
\newblock URL \url{https://arxiv.org/abs/2410.17434}.

\bibitem[Wang et~al.(2025)Wang, He, Hong, Cheng, Zhang, Qi, Gu, Huang, Xu, Dong, Ding, and Tang]{wang2025lvbench}
Weihan Wang, Zehai He, Wenyi Hong, Yean Cheng, Xiaohan Zhang, Ji~Qi, Xiaotao Gu, Shiyu Huang, Bin Xu, Yuxiao Dong, Ming Ding, and Jie Tang.
\newblock Lvbench: An extreme long video understanding benchmark, 2025.
\newblock URL \url{https://arxiv.org/abs/2406.08035}.

\bibitem[Wang et~al.(2026)Wang, Chen, Qin, Wei, Qian, and Bai]{wang2026videohv}
Zheng Wang, Haoran Chen, Haoxuan Qin, Zhipeng Wei, Tianwen Qian, and Cong Bai.
\newblock Think, then verify: A hypothesis-verification multi-agent framework for long video understanding, 2026.
\newblock URL \url{https://arxiv.org/abs/2603.04977}.

\bibitem[Xiao et~al.(2024)Xiao, Tian, Chen, Han, and Lewis]{xiao2024streamingllms}
Guangxuan Xiao, Yuandong Tian, Beidi Chen, Song Han, and Mike Lewis.
\newblock Efficient streaming language models with attention sinks, 2024.
\newblock URL \url{https://arxiv.org/abs/2309.17453}.

\bibitem[Xing et~al.(2025)Xing, Liu, Zhao, Liu, Fu, and Kälviäinen]{xing2025emotionhallucer}
Bohao Xing, Xin Liu, Guoying Zhao, Chengyu Liu, Xiaolan Fu, and Heikki Kälviäinen.
\newblock Emotionhallucer: Evaluating emotion hallucinations in multimodal large language models, 2025.
\newblock URL \url{https://arxiv.org/abs/2505.11405}.

\bibitem[Yang et~al.(2026)Yang, Liu, Guo, Dong, Zhang, Zhang, Wang, Zhou, Xie, Wang, Ouyang, Lin, Cominelli, Cai, Zhang, Zhang, Hong, Widmer, Gringoli, Yang, Li, and Liu]{yang2026egolife}
Jingkang Yang, Shuai Liu, Hongming Guo, Yuhao Dong, Xiamengwei Zhang, Sicheng Zhang, Pengyun Wang, Zitang Zhou, Binzhu Xie, Ziyue Wang, Bei Ouyang, Zhengyu Lin, Marco Cominelli, Zhongang Cai, Yuanhan Zhang, Peiyuan Zhang, Fangzhou Hong, Joerg Widmer, Francesco Gringoli, Lei Yang, Bo~Li, and Ziwei Liu.
\newblock Egolife: Towards egocentric life assistant, 2026.
\newblock URL \url{https://arxiv.org/abs/2503.03803}.

\bibitem[Yeo et~al.(2026)Yeo, Kim, Yoon, and Hwang]{yeo2026worldmm}
Woongyeong Yeo, Kangsan Kim, Jaehong Yoon, and Sung~Ju Hwang.
\newblock Worldmm: Dynamic multimodal memory agent for long video reasoning, 2026.
\newblock URL \url{https://arxiv.org/abs/2512.02425}.

\bibitem[Zhang et~al.(2026)Zhang, Cheng, Deng, Li, Lian, Chen, Liu, Wang, Zhang, Zhang, Guo, Zhu, Wu, Wang, Zheng, Peng, Wu, Wang, Li, Ye, and Heng]{zhang2026mmeemotion}
Fan Zhang, Zebang Cheng, Chong Deng, Haoxuan Li, Zheng Lian, Qian Chen, Huadai Liu, Wen Wang, Yi-Fan Zhang, Renrui Zhang, Ziyu Guo, Zhihong Zhu, Hao Wu, Haixin Wang, Yefeng Zheng, Xiaojiang Peng, Xian Wu, Kun Wang, Xiangang Li, Jieping Ye, and Pheng-Ann Heng.
\newblock Mme-emotion: A holistic evaluation benchmark for emotional intelligence in multimodal large language models, 2026.
\newblock URL \url{https://arxiv.org/abs/2508.09210}.

\bibitem[Zhang et~al.(2025{\natexlab{a}})Zhang, Wang, Tang, Liu, Feng, and Jin]{zhang2025flashvstream}
Haoji Zhang, Yiqin Wang, Yansong Tang, Yong Liu, Jiashi Feng, and Xiaojie Jin.
\newblock Flash-vstream: Efficient real-time understanding for long video streams, 2025{\natexlab{a}}.
\newblock URL \url{https://arxiv.org/abs/2506.23825}.

\bibitem[Zhang et~al.(2025{\natexlab{b}})Zhang, Wang, Zhu, Qin, Wan, Zhang, and Yang]{zhang2025videmo}
Zhicheng Zhang, Weicheng Wang, Yongjie Zhu, Wenyu Qin, Pengfei Wan, Di~Zhang, and Jufeng Yang.
\newblock Videmo: Affective-tree reasoning for emotion-centric video foundation models, 2025{\natexlab{b}}.
\newblock URL \url{https://arxiv.org/abs/2511.02712}.

\bibitem[Zhao et~al.(2025{\natexlab{a}})Zhao, Wei, and Bo]{zhao2025r1omni}
Jiaxing Zhao, Xihan Wei, and Liefeng Bo.
\newblock R1-omni: Explainable omni-multimodal emotion recognition with reinforcement learning, 2025{\natexlab{a}}.
\newblock URL \url{https://arxiv.org/abs/2503.05379}.

\bibitem[Zhao et~al.(2025{\natexlab{b}})Zhao, Yang, Peng, Bai, Yao, Sun, Chen, Fu, chen, Wei, and Bo]{zhao2025humanomni}
Jiaxing Zhao, Qize Yang, Yixing Peng, Detao Bai, Shimin Yao, Boyuan Sun, Xiang Chen, Shenghao Fu, Weixuan chen, Xihan Wei, and Liefeng Bo.
\newblock Humanomni: A large vision-speech language model for human-centric video understanding, 2025{\natexlab{b}}.
\newblock URL \url{https://arxiv.org/abs/2501.15111}.

\bibitem[Zhao et~al.(2026)Zhao, Lin, Li, Zhang, Peng, Zhang, and Wei]{zhao2026pyvisionrl}
Shitian Zhao, Shaoheng Lin, Ming Li, Haoquan Zhang, Wenshuo Peng, Kaipeng Zhang, and Chen Wei.
\newblock Pyvision-rl: Forging open agentic vision models via rl, 2026.
\newblock URL \url{https://arxiv.org/abs/2602.20739}.

\end{thebibliography}
\bibliographystyle{iclr2027_conference}

\clearpage
\appendix

\section*{Appendix Overview}

\begingroup
\setlength{\parindent}{0pt}
\setlength{\parskip}{0pt}
\newcommand{\appoverviewline}[4]{%
  \noindent\hspace*{#1}%
  \hyperref[#3]{\makebox[#2][l]{\ref*{#3}}#4}%
  \nobreak\hspace{0.7em}%
  {\normalfont\color{black!40}\leaders\hbox{.\kern0.35em}\hfill}%
  \nobreak\hspace{0.7em}%
  \hyperref[#3]{\makebox[1.8em][r]{\pageref*{#3}}}\par
}
\newcommand{\appoverviewsection}[2]{%
  \addvspace{0.85\baselineskip}%
  {\bfseries\appoverviewline{0pt}{2.3em}{#1}{#2}}%
}
\newcommand{\appoverviewsubsection}[2]{%
  \vspace{2pt}%
  {\small\appoverviewline{1.4em}{3.0em}{#1}{#2}}%
}
\newcommand{\appoverviewdetail}[2]{%
  \vspace{1pt}%
  {\small\appoverviewline{2.8em}{3.6em}{#1}{#2}}%
}

\appoverviewsection{app:examples}{Dataset Examples}
\appoverviewsubsection{app:examples-scene}{Scene-Level Examples}
\appoverviewsubsection{app:examples-episode}{Episode-Level Examples}
\appoverviewsubsection{app:data-format}{Data Format}

\appoverviewsection{app:emotic}{EMOTIC Emotion Categories}

\appoverviewsection{app:annotation}{Annotation Details}
\appoverviewsubsection{app:annotation-sources}{Video Sources}
\appoverviewsubsection{app:annotation-g1}{G1 Annotation}
\appoverviewsubsection{app:annotation-g2}{G2 Annotation}
\appoverviewsubsection{app:annotation-qc}{Quality Control}

\appoverviewsection{app:evaluation}{Evaluation Protocol}
\appoverviewsubsection{app:evaluation-labels}{Label-based Evaluation}
\appoverviewsubsection{app:evaluation-judge}{Rubric-based Evaluation}
\appoverviewdetail{app:evaluation-rubrics}{Scoring Rubrics}
\appoverviewdetail{app:evaluation-prompt}{Judge Model Prompt}

\appoverviewsection{app:implementation}{Implementation Details}

\appoverviewsection{app:case-studies}{Case Study}

\appoverviewsection{app:longemobench}{Psychological Foundations}
\appoverviewsubsection{app:psych-context}{Context and Temporal Dynamics of Emotion}
\appoverviewsubsection{app:psych-design}{Implications for Benchmark Design}

\appoverviewsection{app:limitations}{Limitations and Ethical Considerations}

\vspace{0.9\baselineskip}
\hrule height 0.5pt
\endgroup
\vfill
\clearpage

\section{Dataset Examples}
\label{app:examples}

We present representative examples of the five Scene-Level (G1) tasks and the three Episode-Level (G2) tasks in LongEmoBench. The examples illustrate the emotional evidence required at each temporal scope and the answer forms used by different tasks. We then describe the shared data format for questions and reference annotations.

\subsection{Scene-Level Examples}
\label{app:examples-scene}

\begin{figure}[!htbp]
    \centering
    \includegraphics[width=\linewidth]{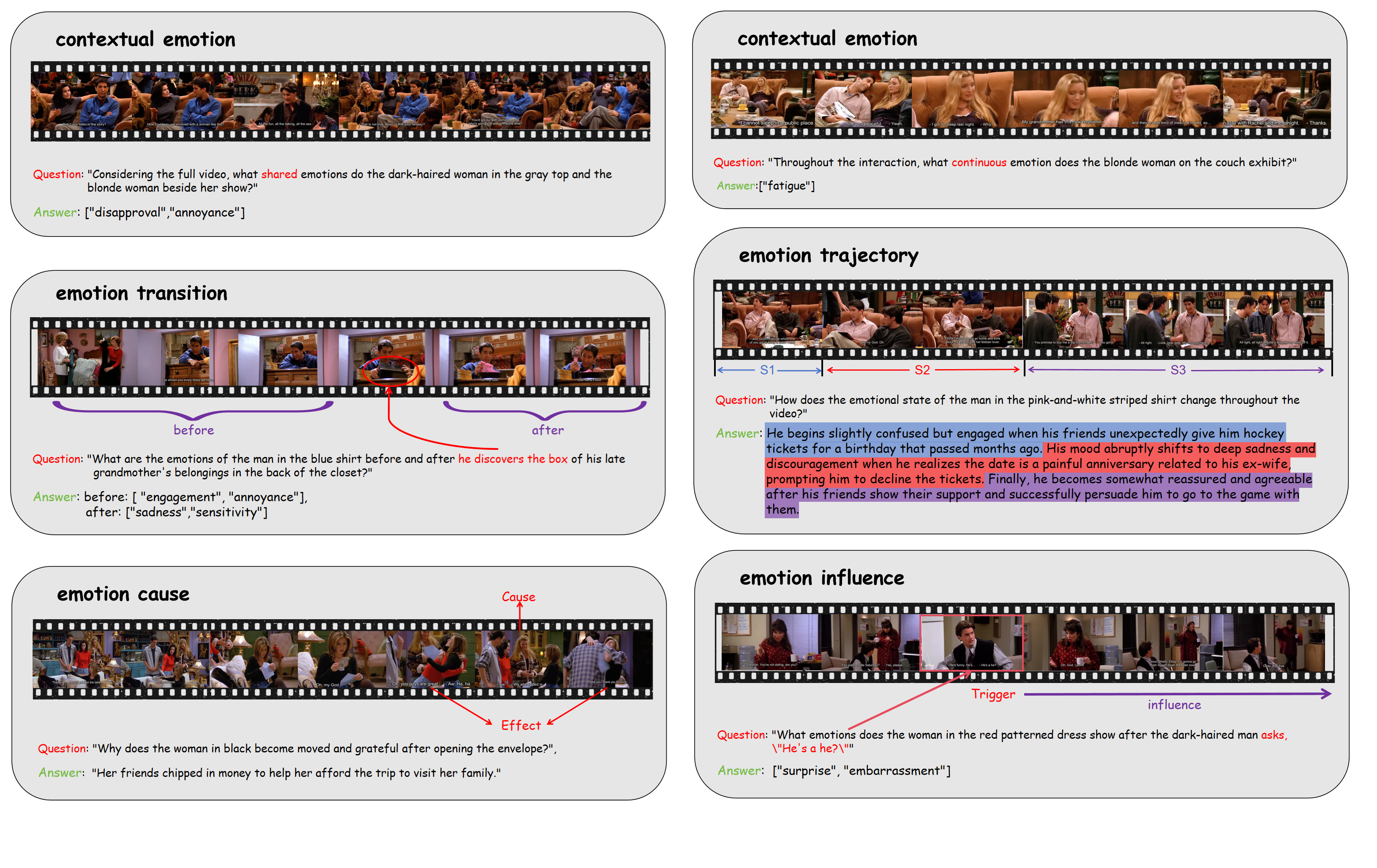}
    \caption{\textbf{Scene-Level (G1) examples.} Six panels illustrate five tasks, including two forms of Contextual Emotion. Temporal brackets, colored stages, and event annotations highlight the distinctions between identifying states, tracking changes, and explaining or identifying emotional responses.}
    \label{fig:app-g1-examples}
\end{figure}

Figure~\ref{fig:app-g1-examples} illustrates all five G1 tasks. The two \textit{Contextual Emotion} examples distinguish shared emotions from an individual state sustained throughout an interaction: the former identifies shared disapproval and annoyance, whereas the latter identifies continuing fatigue. In the latter example, although the target woman briefly exhibits aversion, this transient reaction is excluded from the reference answer because the question asks for emotions that persist throughout the interaction. \textit{Emotion Transition} compares the target person's emotions before and after the discovery of his late grandmother's belongings. \textit{Emotional Trajectory} instead reconstructs successive stages: the man receiving hockey tickets moves from confusion and engagement to sadness associated with a painful anniversary, and finally becomes reassured by his friends. The remaining examples pose two complementary questions about events and emotions. \textit{Emotion Cause} asks why the woman is moved and grateful, requiring the information that her friends have contributed money for her trip. \textit{Emotion Influence} asks for the emotional response to a specified remark, with surprise and embarrassment as the reference labels. These examples require integrating sustained states, event boundaries, or ordered emotional developments within a scene.

\clearpage
\subsection{Episode-Level Examples}
\label{app:examples-episode}

\begin{figure}[!htbp]
    \centering
    \includegraphics[width=\linewidth]{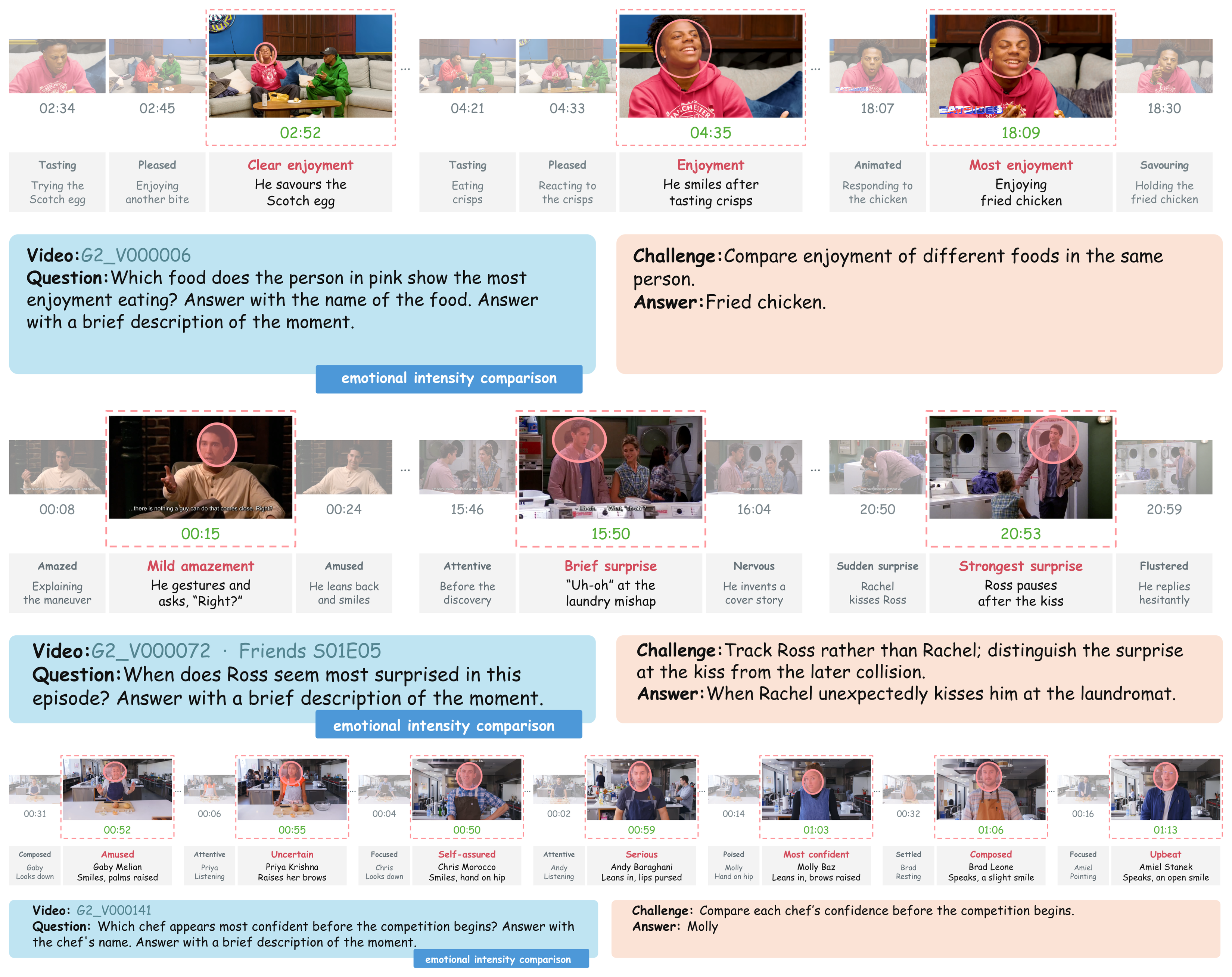}
    \caption{\textbf{Episode-Level emotional intensity comparisons.} The three examples illustrate emotional intensity comparisons across different targets, moments, and people, respectively. Highlighted observations identify the target reactions considered in each comparison.}
    \label{fig:app-g2-intensity}
\end{figure}

\paragraph{Emotional Intensity Comparison.}
Figure~\ref{fig:app-g2-intensity} illustrates comparisons across targets, moments, and people. In the food-tasting video, the model must compare one person's enjoyment of different foods and identify fried chicken. In the episode example, it must compare Ross's surprise across separated events and locate Rachel's unexpected kiss at the laundromat. In the cooking video, it must compare the chefs' confidence within the specified pre-competition period and identify Molly. Each question requires establishing the relevant comparison set and tracking the correct person across observations.

\begin{figure}[!htbp]
    \centering
    \includegraphics[width=\linewidth,height=0.88\textheight,keepaspectratio]{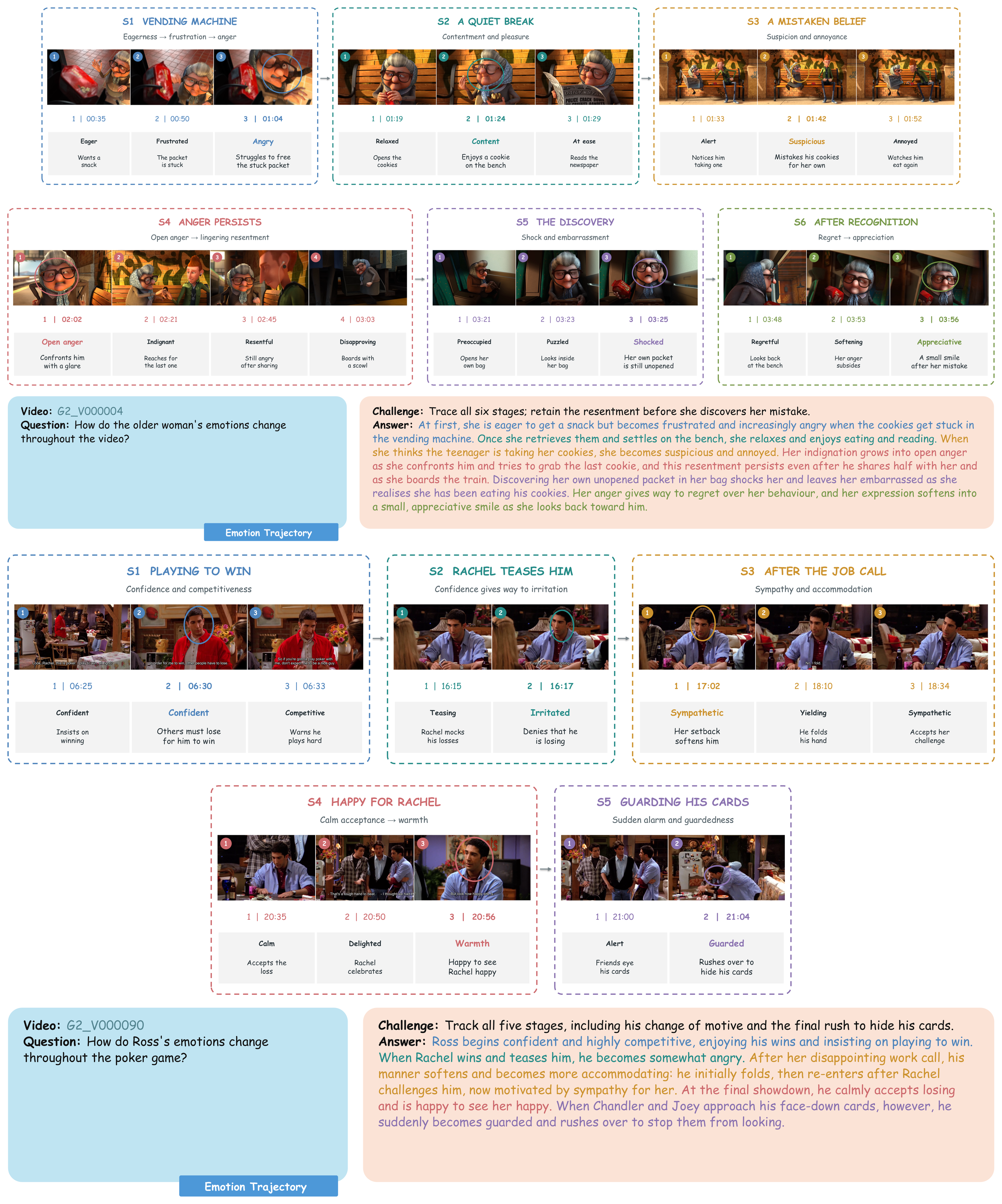}
    \caption{\textbf{Episode-Level emotional trajectories.} The examples organize an animated narrative into six stages and Ross's poker storyline into five stages. The annotations preserve emotional order, persistence, and turning points.}
    \label{fig:app-g2-trajectory}
\end{figure}

\paragraph{Emotional Trajectory.}
Figure~\ref{fig:app-g2-trajectory} shows how an extended trajectory preserves intermediate stages and their relationships. The older woman's story includes frustration at a vending machine, a contented break, suspicion and anger over the cookies, lingering resentment, shock at discovering her mistake, and subsequent regret and appreciation. In the poker storyline, Ross progresses from confidence and competitiveness through irritation, sympathy after Rachel's disappointing job call, and happiness for her, before becoming guarded when his friends approach his cards. Correct answers retain these developments, including persistent emotions and late changes that a beginning-to-end summary would omit.

\begin{figure}[!htbp]
    \centering
    \includegraphics[width=\linewidth]{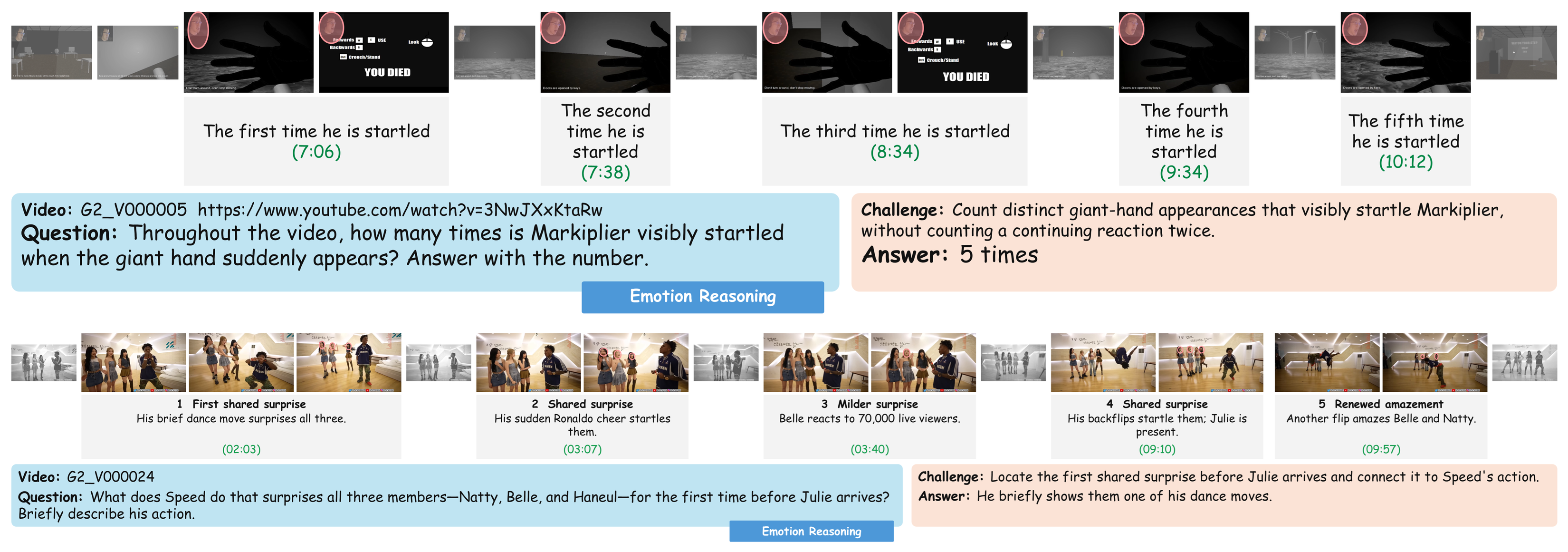}
    \caption{\textbf{Episode-Level emotional reasoning with definite answers.}}
    \label{fig:app-g2-reasoning}
\end{figure}

\clearpage
\paragraph{Emotional Reasoning.}
Figure~\ref{fig:app-g2-reasoning} illustrates reasoning with definite answers. The first question asks how many distinct appearances of a giant hand visibly startle the streamer; the answer is five, requiring reaction detection and aggregation without counting a continuing reaction twice. The second asks which action first surprises all three specified participants before another participant arrives; the answer is a brief dance move. This requires jointly resolving the action, the shared emotional response, and the temporal condition. Both examples depend on emotional evidence across the video despite their short answers.

\begin{figure}[!htbp]
    \centering
    \includegraphics[width=\linewidth]{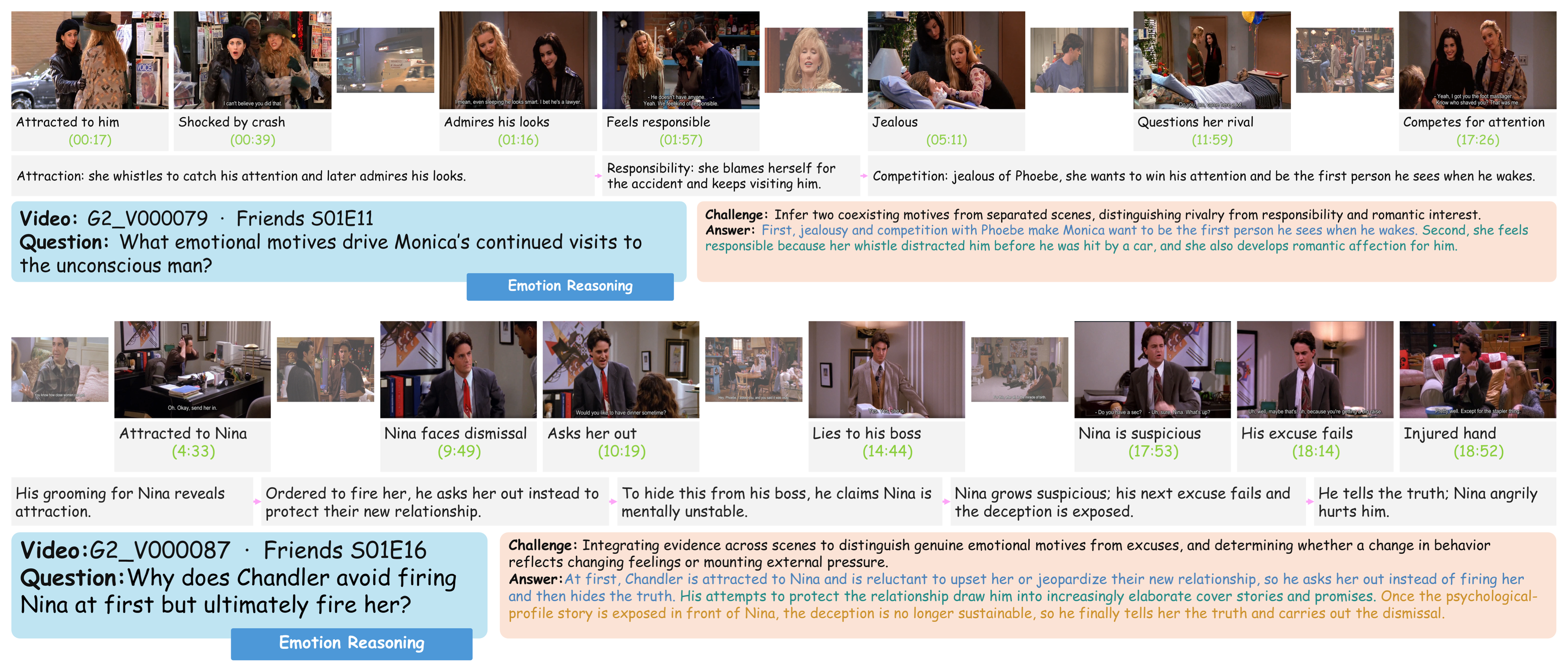}
    \caption{\textbf{Episode-Level causal and motivational reasoning.}}
    \label{fig:app-g2-motives}
\end{figure}

Figure~\ref{fig:app-g2-motives} illustrates causal and motivational reasoning. Monica's repeated hospital visits reflect several coexisting motives: rivalry with Phoebe, a sense of responsibility for the accident, and romantic interest. An answer must connect these motives to evidence from separated scenes, rather than attribute every visit to the same immediate trigger. The second example asks why Chandler initially avoids firing Nina but eventually does so. The reference explanation connects his attraction and efforts to protect the relationship with his escalating cover stories and their eventual exposure. The challenge is to distinguish his emotional motives from his excuses and to explain why his behavior changes as the deception becomes unsustainable.

\clearpage
\subsection{Data Format}
\label{app:data-format}

Each question is stored as a JSON record with the ten top-level fields in Table~\ref{tab:app-data-fields}.

\begin{table}[!htbp]
    \centering
    \small
    \renewcommand{\arraystretch}{1.17}
    \caption{\textbf{Shared JSON fields for G1 and G2.} Nested fields and the use of \texttt{null} are described in the corresponding entries.}
    \label{tab:app-data-fields}
    \begin{tabularx}{\linewidth}{@{}lX@{}}
        \toprule
        \textbf{Field} & \textbf{Content} \\
        \midrule
        \texttt{question\_id} & Stable question identifier within its granularity. \\
        \texttt{video\_id} & Identifier of the corresponding prepared video. \\
        \texttt{source} & Object with \texttt{from}, the original URL or series/season/episode identifier, and \texttt{segments}, an ordered list of \texttt{\{start, end\}} intervals in seconds on the original video timeline. Intervals follow concatenation order; \texttt{segments} is \texttt{null} when the complete original video is used. \\
        \texttt{granularity} & \texttt{clip} for G1 or \texttt{episode} for G2. \\
        \texttt{type} & Task-category string, e.g., \texttt{contextual emotion}, \texttt{emotion trajectory}, or \texttt{emotional reasoning}. \\
        \texttt{question} & Open-ended question text, including any requested answer format or temporal scope. \\
        \texttt{answer} & Reference answer: an emotion-label array, a before/after object, or a natural-language string, depending on the task. \\
        \texttt{answer\_details} & Object with \texttt{description}, explaining the content and use of reference items, and \texttt{items}, storing them; \texttt{null} when additional structure is unnecessary. \\
        \texttt{rubric} & Object with the overall \texttt{criterion} and \texttt{scores}, a mapping from score values to textual criteria; \texttt{null} for label-based tasks. \\
        \texttt{subtitles} & Full subtitle list, or \texttt{null} when not included. Entries contain \texttt{id} (\texttt{U1}, \texttt{U2}, \ldots\ in video order), \texttt{t} (a \texttt{[start, end]} array in original-video seconds), \texttt{speaker}, and \texttt{text}. \\
        \bottomrule
    \end{tabularx}
\end{table}

The following records illustrate a G1 emotion transition and three G2 tasks with different scoring scales.

\lstdefinestyle{appjson}{
    basicstyle=\ttfamily\footnotesize,
    columns=fullflexible,
    keepspaces=true,
    showstringspaces=false,
    breaklines=true,
    breakatwhitespace=true,
    breakindent=1em,
    frame=single,
    rulecolor=\color{black!25},
    backgroundcolor=\color{black!3},
    framesep=5pt,
    xleftmargin=6pt,
    xrightmargin=6pt,
    aboveskip=0.5\baselineskip,
    belowskip=0.5\baselineskip,
    literate={’}{{\textquoteright}}1
}

\paragraph{G1: emotion transition.} The record corresponds to Figure~\ref{fig:app-g1-examples}.

\begin{lstlisting}[style=appjson]
{
  "question_id": "G1_Q000126",
  "video_id": "G1_V000116",
  "source": {"from": "Friends S01E08", "segments": [{"start": 687.6, "end": 778}]},
  "granularity": "clip",
  "type": "emotion transition",
  "question": "What are the emotions of the man in the blue shirt before and after he discovers the box of his late grandmother's belongings in the back of the closet?",
  "answer": {
    "before": ["engagement", "annoyance"],
    "after": ["sadness", "sensitivity"]
  },
  "answer_details": null,
  "rubric": null,
  "subtitles": [
    {"id": "U1", "t": [687.61, 688.82], "speaker": "Ross", "text": "This one?"},
    ...
  ]
}
\end{lstlisting}

\clearpage
\paragraph{G2: intensity comparison (0--1).} The record corresponds to Figure~\ref{fig:app-g2-intensity}.

\begin{lstlisting}[style=appjson]
{
  "question_id": "G2_Q000013",
  "video_id": "G2_V000006",
  "source": {"from": "https://www.youtube.com/watch?v=4Ho4RJsIdvM", "segments": null},
  "granularity": "episode",
  "type": "emotional intensity comparison",
  "question": "Which food does the person in pink show the most enjoyment eating? Answer with the name of the food.",
  "answer": "Fried chicken.",
  "answer_details": null,
  "rubric": {...},
  "subtitles": null
}
\end{lstlisting}

\paragraph{G2: emotion trajectory (0--4).} The record corresponds to Figure~\ref{fig:app-g2-trajectory}.

\begin{lstlisting}[style=appjson]
{
  "question_id": "G2_Q000327",
  "video_id": "G2_V000090",
  "source": {"from": "Friends S01E18", "segments": null},
  "granularity": "episode",
  "type": "emotion trajectory",
  "question": "How do Ross's emotions change throughout the poker game?",
  "answer": "Ross begins confident and highly competitive, enjoying his wins and insisting on playing to win. When Rachel wins and teases him, he becomes somewhat angry. After her disappointing work call, his manner softens and becomes more accommodating: he initially folds, then re-enters after Rachel challenges him, now motivated by sympathy for her. At the final showdown, he calmly accepts losing and is happy to see her happy. When Chandler and Joey approach his face-down cards, however, he suddenly becomes guarded and rushes over to stop them from looking.",
  "answer_details": {
    "description": "The required stages of the emotional trajectory, listed in chronological order. Assess the emotions at each stage and how they develop across stages.",
    "items": [
      {"id": "S1",
       "description": "Ross begins confident and highly competitive, enjoying his wins and insisting on playing to win.",
       "anchor": "The early poker games, rematch challenges and his play-to-win speech.",
       "emotion": "confidence, enjoyment, competitiveness",
       "intensity": null
      },
      {"id": "S2",
       "description": "When Rachel wins and teases him, Ross becomes somewhat angry rather than remaining comfortably in control.",
       "anchor": "Rachel wins with four sixes, raises against him and teases him about losing.",
       "emotion": "anger, competitiveness",
       "intensity": null
      },
      {"id": "S3",
       "description": "After Rachel receives the disappointing work call, Ross softens and becomes more accommodating: he initially folds, then re-enters after she challenges him, now motivated by sympathy for her.",
       "anchor": "The rejection call, his initial fold and Rachel challenging him to continue.",
       "emotion": "sympathy",
       "intensity": null
      },
      {"id": "S4",
       "description": "At the final showdown, Ross calmly accepts losing and is happy to see Rachel happy rather than reacting with his earlier anger.",
       "anchor": "The final showdown, Rachel celebrating and his remark about how happy she is.",
       "emotion": "calmness, happiness, warmth",
       "intensity": null
      },
      {"id": "S5",
       "description": "When Chandler and Joey approach his face-down cards, Ross suddenly becomes guarded and rushes over to stop them from looking.",
       "anchor": "The silent action after the conversation about Rachel being happy, before the Pictionary scene.",
       "emotion": "alarm, guardedness",
       "intensity": null
      }
    ]
  },
  "rubric": {...},
  "subtitles": null
}
\end{lstlisting}

\paragraph{G2: emotional reasoning (0--3).} The record corresponds to Figure~\ref{fig:app-g2-motives}.

\begin{lstlisting}[style=appjson]
{
  "question_id": "G2_Q000279",
  "video_id": "G2_V000079",
  "source": {"from": "Friends S01E11", "segments": null},
  "granularity": "episode",
  "type": "emotional reasoning",
  "question": "What emotional motives drive Monica’s continued visits to the unconscious man?",
  "answer": "First, jealousy and competition with Phoebe make Monica want to be the first person he sees when he wakes. Second, she feels responsible because her whistle distracted him before he was hit by a car, and she also develops romantic affection for him.",
  "answer_details": {
    "description": "The necessary causal factors that jointly form the reference explanation. Consider the items together when assessing whether the explanation is complete.",
    "items": [
      {"id": "R1",
       "description": "Monica’s growing rivalry with Phoebe makes her want to win his attention and be the first person he sees when he wakes."
      },
      {"id": "R2",
       "description": "Monica feels responsible because her whistle distracted him before he was hit by a car, and she also develops romantic affection for him."
      }
    ]
  },
  "rubric": {...},
  "subtitles": null
}
\end{lstlisting}

\section{EMOTIC Emotion Categories}
\label{app:emotic}

We adopt the 26-category EMOTIC vocabulary~\citep{Kosti_2019}\footnote{ \url{https://s3.sunai.uoc.edu/emotic/annotations.html}.} for Contextual Emotion, Emotion Transition, and Emotion Influence. Designed for emotion recognition in context, it includes fine-grained states such as anticipation, engagement, and sympathy that are relevant to character interactions. Multiple labels can represent co-occurring emotions, while the shared vocabulary standardizes annotation targets and enables consistent label-based evaluation across videos and models.

\begin{center}
\begingroup
\setlength{\fboxsep}{8pt}
\setlength{\fboxrule}{0.4pt}
\fcolorbox{black!25}{black!3}{%
\begin{minipage}{\dimexpr\linewidth-2\fboxsep-2\fboxrule\relax}
\small
\textbf{EMOTIC: 26 emotion categories}\par
\emph{Official category definitions.}\par\smallskip
\renewcommand{\arraystretch}{1.12}
\begin{tabularx}{\linewidth}{@{}>{\bfseries}l X@{}}
1. Peace: & well being and relaxed; no worry; having positive thoughts or sensations; satisfied. \\
2. Affection: & fond feelings; love; tenderness \\
3. Esteem: & feelings of favorable opinion or judgment; respect; admiration; gratefulness \\
4. Anticipation: & state of looking forward; hoping on or getting prepared for possible future events \\
5. Engagement: & paying attention to something; absorbed into something; curious; interested \\
6. Confidence: & feeling of being certain; conviction that an outcome will be favorable; encouraged; proud \\
7. Happiness: & feeling delighted; feeling enjoyment or amusement \\
8. Pleasure: & feeling of delight in the senses \\
9. Excitement: & feeling enthusiasm; stimulated; energetic \\
10. Surprise: & sudden discovery of something unexpected \\
11. Sympathy: & state of sharing others’ emotions, goals or troubles; supportive; compassionate \\
12. Doubt/Confusion: & difficulty to understand or decide; thinking about different options \\
13. Disconnection: & feeling not interested in the main event of the surrounding; indifferent; bored; distracted \\
14. Fatigue: & weariness; tiredness; sleepy \\
15. Embarrassment: & feeling ashamed or guilty \\
16. Yearning: & strong desire to have something; jealous; envious; lust \\
17. Disapproval: & feeling that something is wrong or reprehensible; contempt; hostile \\
18. Aversion: & feeling disgust, dislike, repulsion; feeling hate \\
19. Annoyance: & bothered by something or someone; irritated; impatient; frustrated \\
20. Anger: & intense displeasure or rage; furious; resentful \\
21. Sensitivity: & feeling of being physically or emotionally wounded; feeling delicate or vulnerable \\
22. Sadness: & feeling unhappy, sorrow, disappointed, or discouraged \\
23. Disquietment: & nervous; worried; upset; anxious; tense; pressured; alarmed \\
24. Fear: & feeling suspicious or afraid of danger, threat, evil or pain; horror \\
25. Pain: & physical suffering \\
26. Suffering: & psychological or emotional pain; distressed; anguished \\
\end{tabularx}
\end{minipage}%
}
\endgroup
\end{center}

\section{Annotation Details}
\label{app:annotation}

\subsection{Video Sources}
\label{app:annotation-sources}

Video selection for LongEmoBench focuses on the context, emotional developments, and audiovisual evidence needed for emotion understanding. We prioritize videos with rich emotional interactions and coherent narrative contexts, enabling models to interpret emotions through characters' experiences, relationships, and immediate reactions. For long videos, we emphasize connections between events, changes in characters' emotional states, and emotional cues distributed over time. These properties support questions about emotional trajectories, intensity comparisons, and emotional reasoning.

We draw materials from sitcoms, such as \textit{Friends} and \textit{Modern Family}, and online videos covering interviews, reality shows, and gameplay, capturing diverse interpersonal relationships, interaction styles, and emotional expressions. Their unfolding events provide context for how emotions arise and change. For example, a character may initially resist living with family, reluctantly accept the arrangement, then experience pressure and frustration during daily interactions before finding relief when a solution emerges. Such situations allow questions to examine emotional changes, differences in intensity, and their causes, requiring models to connect evidence from different points in the video.

\subsection{G1 Annotation}
\label{app:annotation-g1}

\begin{figure}[!htbp]
    \centering
    \includegraphics[width=\linewidth]{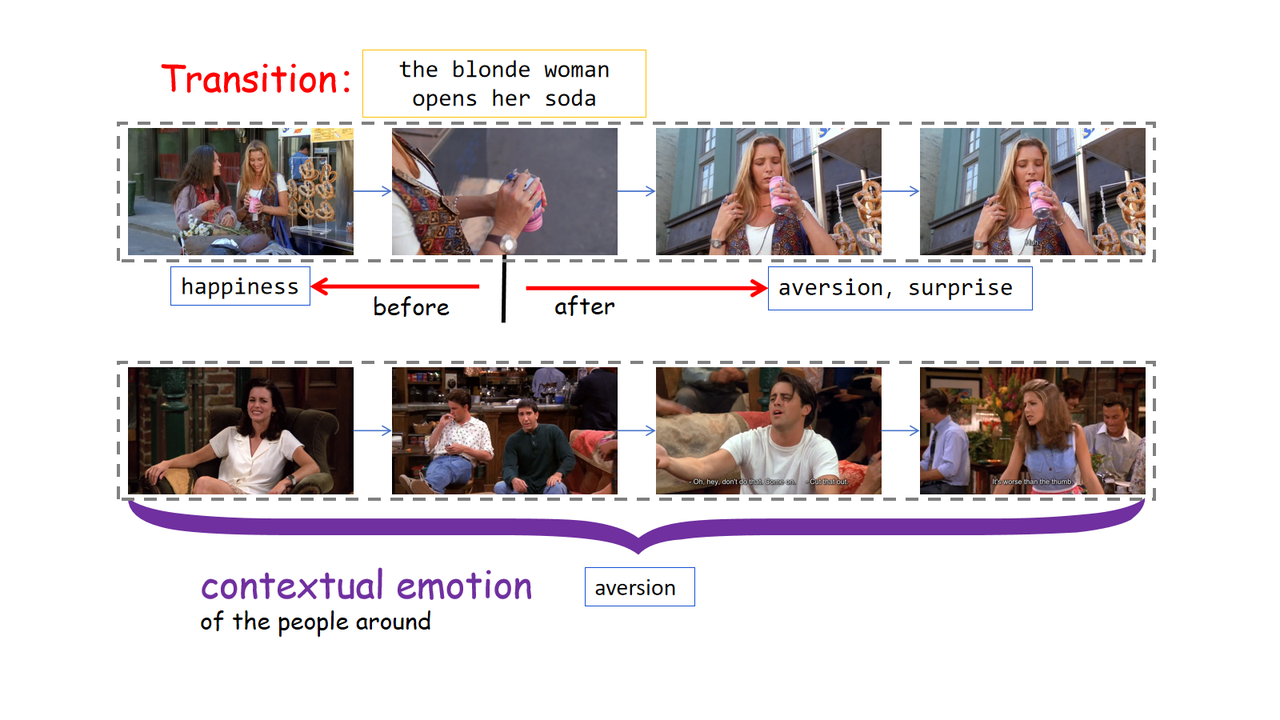}
    \caption{\textbf{Annotating contextual emotions and event-bounded transitions.} The upper sequence separates the woman's emotions before and after opening her soda. The lower sequence identifies aversion shared by the surrounding people during the interaction.}
    \label{fig:app-annotation-states}
\end{figure}

\begin{figure}[!htbp]
    \centering
    \includegraphics[width=\linewidth]{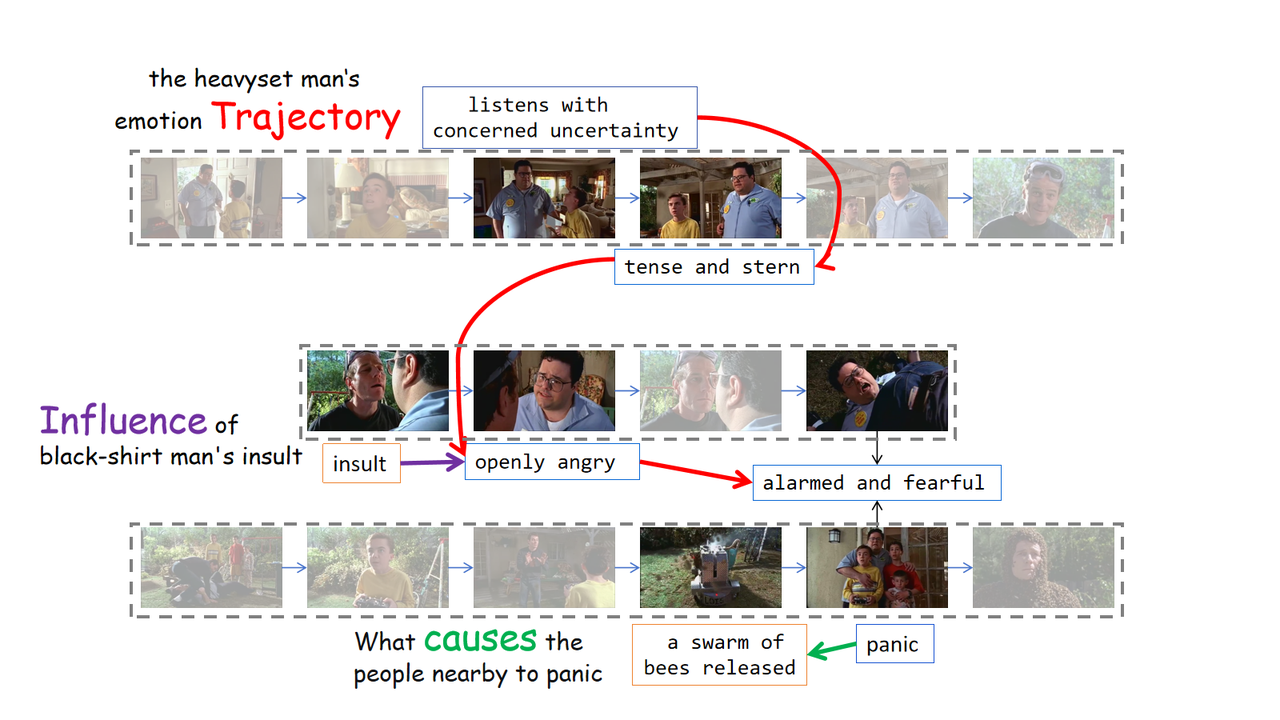}
    \caption{\textbf{Annotating trajectories, influences, and causes.} The connected states trace the target man's emotional development. The insult anchors an Emotion Influence question, while the release of bees provides the explanation for the group's panic in an Emotion Cause question.}
    \label{fig:app-annotation-process}
\end{figure}

\paragraph{Model-assisted annotation workflow.}
G1 combines model assistance with human annotation. Starting from the videos described in Section~\ref{app:annotation-sources}, we use \texttt{gpt-5.6-sol} to identify temporal boundaries and prepare candidate clips. We then construct draft questions and reference answers for five task types: Contextual Emotion, Emotion Transition, Emotion Trajectory, Emotion Cause, and Emotion Influence. Human annotators inspect the clips and revise or rewrite the drafts according to the guidelines below, checking the task assignment, question scope, and reference answer against the video. Figures~\ref{fig:app-annotation-states} and~\ref{fig:app-annotation-process} illustrate the distinctions used during question construction and revision.

\paragraph{Annotation guidelines.}
\begin{itemize}
    \item \textbf{Clip boundaries and question scope.} Retain the smallest complete context needed to answer the question, from before the first necessary cue until the relevant response or outcome is clear. Give each question one target, identify people by observable attributes, and use neutral event descriptions without revealing the answer or supplying unsupported emotional or causal conclusions.
    \item \textbf{Contextual Emotion.} Identify persistent individual emotions, emotions shared by the specified people, or the group's overall emotional tone, as requested. Exclude transient reactions from persistent-emotion answers; persistence does not require an identical expression in every frame. Record an EMOTIC label set (Figure~\ref{fig:app-annotation-states}, lower sequence).
    \item \textbf{Emotion Transition.} Specify an observable event, utterance, action, or interaction and retain evidence on both sides. Record separate \texttt{before} and \texttt{after} EMOTIC label sets. The emotion categories must change; a change in intensity alone belongs to Emotion Trajectory (Figure~\ref{fig:app-annotation-states}, upper sequence).
    \item \textbf{Emotion Trajectory.} Describe all necessary emotional stages and their relationships in chronological order, including intensity changes when requested and supported. Divide stages by meaningful emotional developments, not by shots or utterances; similar repeated reactions may form one recurrent stage. Record the necessary stages in \texttt{answer\_details}.
    \item \textbf{Emotion Cause.} Explain the video-supported causes, motives, or intentions and their connection to the target emotion or decision. Temporal succession alone does not establish causation. Record necessary explanatory factors in \texttt{answer\_details} when needed, without inventing psychological interpretations.
    \item \textbf{Emotion Influence.} Identify the EMOTIC labels of the target's actual response to a specified person, utterance, event, or interaction. Verify that the video supports the influence: an attempt to comfort or encourage someone does not establish its emotional effect. Intensity-only changes belong to Emotion Trajectory. Figure~\ref{fig:app-annotation-process} illustrates the distinction between trajectory, cause, and influence.
    \item \textbf{Revision and answer verification.} Check answers using expressions, actions, prosody, dialogue, and context. When converting an existing question to open-ended form, recheck its full answer scope instead of merely deleting options. Keep only the information needed to answer the question, accept equivalent wording for textual answers, and exclude unsupported or unresolved drafts.
\end{itemize}

\clearpage
\subsection{G2 Annotation}
\label{app:annotation-g2}

All G2 questions are authored from scratch by human annotators after they watch the complete video. Annotators identify emotional interactions and evidence distributed across the video, then formulate questions and reference answers following the guidelines below. Table~\ref{tab:app-g2-annotation-guidelines} summarizes the requirements and example questions for the three task types.
\begin{itemize}
    \item \textbf{Selecting the target and scope.} Select a character, interaction, or storyline with emotion-relevant evidence across the video. Decide whether to ask about an intensity comparison, an emotional trajectory, or an emotion-related inference, and define the people, events, and temporal scope needed for that question.
    \item \textbf{Evidence across time.} Design questions that require connecting information from different parts of the video. Each relevant interval must contribute to the answer. Favor questions whose key judgments require expressions, actions, tone of voice, or other audiovisual evidence beyond subtitles, together with context.
    \item \textbf{Formulating the question.} Write a natural, open-ended question around the selected target. Use clear references and neutral context to establish the scope, leaving the emotional conclusion, comparison result, or explanation to be inferred from the video without revealing the answer location.
    \item \textbf{Writing the reference answer.} Derive the answer from the relevant audiovisual evidence and express the requested result or necessary explanation, allowing equivalent wording. Comparisons identify a distinguishable moment, target, or intensity relation. Trajectories preserve the necessary stages and their chronological relationships, grouping repeated similar reactions when appropriate. Reasoning answers state a definite result or explain the necessary causes, motives, or intentions. Add structured details for required stages or explanatory factors when needed, without unsupported psychological assumptions.
\end{itemize}

\begin{table}[!htbp]
    \centering
    \small
    \setlength{\tabcolsep}{4pt}
    \renewcommand{\arraystretch}{1.15}
    \renewcommand{\tabularxcolumn}[1]{m{#1}}
    \caption{Task-specific question construction guidelines for G2 annotation.}
    \label{tab:app-g2-annotation-guidelines}
    \begin{tabularx}{\linewidth}{@{}>{\raggedright\arraybackslash}m{0.17\linewidth}| >{\raggedright\arraybackslash}X@{}}
        \toprule
        \textbf{Task} & \textbf{Requirements and Examples} \\
        \midrule
        Emotional intensity comparison & Specify the emotion and comparison scope. Ask about the strongest moment or person, or another supported intensity relation, without listing candidate moments. Judge intensity from audiovisual evidence and context, not event severity or action magnitude. Pose extremum questions only when a unique answer is supported.\par\smallskip\textbf{Example:} When does Ross seem most surprised in this episode? Answer with a brief description of the moment. \\
        \midrule
        Emotional trajectory & Specify the person and storyline; narrow the scope when multiple storylines interleave. Ask about emotional development or changes in the intensity of one emotion, without listing stages or revealing turning points. Preserve the necessary process rather than reducing it to its endpoints.\par\smallskip\textbf{Example:} How do Ross's emotions change throughout the poker game? \\
        \midrule
        Emotional reasoning & Require an emotional judgment about a state, attitude, cause, motive, intention, or outcome, beyond plot retrieval. Distinguish facts from characters' beliefs and stated excuses; temporal order alone does not establish psychological causation. For counts, define the unit and count multiple shots of one reaction only once.\par\smallskip\textbf{Example:} What emotional motives drive Monica's continued visits to the unconscious man? \\
        \bottomrule
    \end{tabularx}
\end{table}

\subsection{Quality Control}
\label{app:annotation-qc}

Quality review is conducted by eight experienced AI practitioners. Before formal annotation, participants complete five hours of annotation training. After annotation, two reviewers independently assess each question and its reference answer. Items with disagreements are returned to the original annotator for re-annotation and renewed independent review. An item is retained only when both reviewers agree to accept it; items for which agreement cannot be reached are removed.

\paragraph{Inter-reviewer agreement.}
For the reviewers' initial review decisions, the observed agreement rate is 84.86\% for G1 and 71.25\% for G2, with nominal Krippendorff's $\alpha$ values~\citep{krippendorff2011alpha} of 0.133 and 0.118, respectively.

\section{Evaluation Protocol}
\label{app:evaluation}

We evaluate G1 label predictions with set-based metrics and use a shared LLM judge for the remaining G1 and G2 tasks. This section specifies the metrics, structured references, scoring rubrics, and judge prompt.

\subsection{Label-based Evaluation}
\label{app:evaluation-labels}

For G1 Contextual Emotion and Emotion Influence, let $Y$ and $\widehat{Y}$ denote the reference and predicted EMOTIC label sets for a question. Labels are normalized for case and whitespace and deduplicated before scoring. F1 is computed as
\begin{equation}
 F_1=\frac{2|Y\cap\widehat{Y}|}{|Y|+|\widehat{Y}|}.
 \label{eq:app-label-f1}
\end{equation}
An empty predicted set receives zero F1, and labels outside the vocabulary count as incorrect predictions. For Emotion Transition, the \texttt{before} and \texttt{after} sets are scored separately and averaged:
\begin{equation}
 F_1^{\mathrm{transition}}
 =\frac{F_1^{\mathrm{before}}+F_1^{\mathrm{after}}}{2}.
 \label{eq:app-transition-f1}
\end{equation}
The two sides are weighted equally. If only one side is provided, the question score is half of that side's F1. Task-level F1 is the average of the question-level scores.

\subsection{Rubric-based Evaluation}
\label{app:evaluation-judge}

We use DeepSeek-V4-Flash~\citep{deepseekai2026deepseekv4} as the LLM judge. Three rubric families cover the tasks that require model-based evaluation:
\begin{itemize}
 \item \textbf{Result correctness:} G2 Emotional Intensity Comparison and definite-result questions in G2 Emotional Reasoning, including questions asking for a person, moment, count, or yes/no judgment.
 \item \textbf{Trajectory reconstruction:} Emotional Trajectory in both G1 and G2.
 \item \textbf{Causal explanation:} G1 Emotion Cause and G2 Emotional Reasoning questions asking for causes, motives, or intentions.
\end{itemize}
The rubric follows the question's requirements, independently of the length or format of the model response. Each question stores its complete \texttt{rubric}, with a \texttt{criterion} and a \texttt{scores} mapping from score levels to their descriptions. The same family of rubric is used across granularities wherever applicable.

\paragraph{Reference Answer.}
The \texttt{answer} field gives a complete reference response. When additional structure is useful, \texttt{answer\_details} contains a \texttt{description} and an \texttt{items} list. Its top-level \texttt{description} specifies the required content and how the items should be considered together. An item's \texttt{description} specifies its substantive reference content. The judge uses this information to assess correctness and completeness while accepting semantically equivalent formulations.
\begin{itemize}
 \item \textbf{Trajectory stages:} Chronologically ordered \texttt{S1}, \texttt{S2}, \ldots\ contain \texttt{id}, \texttt{description}, \texttt{anchor}, \texttt{emotion}, and \texttt{intensity}. Descriptions specify the required emotional stages, developments, and recurring patterns. An \texttt{anchor} aligns a stage with an event, utterance, action, or relative position; it need not be a cause. Anchors receive no separate credit and need not be repeated when stage alignment is clear. The \texttt{emotion} field records the stage's emotion terms, while \texttt{intensity} specifies intensity or its change only when required by the question and supported by the video; otherwise it is \texttt{null}.
 \item \textbf{Explanatory factors:} \texttt{R1}, \texttt{R2}, \ldots\ contain \texttt{id} and \texttt{description}, identifying the necessary causes, motives, or intentions to consider together when assessing completeness.
 \item \textbf{Result components:} \texttt{C1}, \texttt{C2}, \ldots\ may specify comparison or result components that must be checked together.
\end{itemize}
These items guide one holistic judgment. The field is \texttt{null} when no supplementary reference information is needed. Reference answers, structured details, and rubrics are reserved for evaluation and are excluded from the tested model's input.

\subsubsection{Scoring Rubrics}
\label{app:evaluation-rubrics}

Tables~\ref{tab:app-rubric-binary}--\ref{tab:app-rubric-explanation} present the scoring criteria applied to the benchmark.

\begingroup
\setlength{\intextsep}{6pt}
\setlength{\textfloatsep}{8pt}
\setlength{\floatsep}{8pt}
\begin{table}[!htbp]
\centering
\small
\caption{Result correctness rubric for G2 Emotional Intensity Comparison and definite-result Emotional Reasoning.}
\label{tab:app-rubric-binary}
\begingroup
\setlength{\tabcolsep}{8pt}
\renewcommand{\arraystretch}{1.18}
\setlength{\extrarowheight}{2pt}
\setlength{\aboverulesep}{0pt}
\setlength{\belowrulesep}{0pt}
\renewcommand{\tabularxcolumn}[1]{m{#1}}
\begin{tabularx}{\linewidth}{>{\centering\arraybackslash}m{0.11\linewidth}|>{\raggedright\arraybackslash}X}
\toprule
\rowcolor{black!6}
 & \textbf{Description} \\[3pt]
\midrule
\textbf{Criterion} & Assess whether the response correctly answers the question. Accept equivalent wording. When answer\_details are supplied, follow their description to determine which items the answer must satisfy. \\[5pt]
\midrule
\textbf{Score} &
\hangindent=1.7em\hangafter=1\noindent\makebox[1.7em][l]{\textbf{1}}The response correctly and completely answers the question, with no errors that affect its correctness.\par\smallskip
\hangindent=1.7em\hangafter=1\noindent\makebox[1.7em][l]{\textbf{0}}The response is incorrect or empty, omits information required by the question, or gives mutually contradictory answers. \\[5pt]
\bottomrule
\end{tabularx}
\endgroup
\end{table}

\begin{table}[!htbp]
\centering
\small
\caption{Trajectory reconstruction rubric, shared by G1 and G2 Emotional Trajectory.}
\label{tab:app-rubric-trajectory}
\begingroup
\setlength{\tabcolsep}{8pt}
\renewcommand{\arraystretch}{1.18}
\setlength{\extrarowheight}{2pt}
\setlength{\aboverulesep}{0pt}
\setlength{\belowrulesep}{0pt}
\renewcommand{\tabularxcolumn}[1]{m{#1}}
\begin{tabularx}{\linewidth}{>{\centering\arraybackslash}m{0.11\linewidth}|>{\raggedright\arraybackslash}X}
\toprule
\rowcolor{black!6}
 & \textbf{Description} \\[3pt]
\midrule
\textbf{Criterion} & Assess the accuracy and completeness of the required emotional stages and their relationships, including intensity changes and recurring patterns where relevant. Assign one holistic score from 0 to 4 using the reference answer and stage annotations. Anchors support stage alignment and receive no separate credit; repeating them is not required. \\[5pt]
\midrule
\textbf{Score} &
\hangindent=1.7em\hangafter=1\noindent\makebox[1.7em][l]{\textbf{4}}The response accurately and completely reconstructs the required emotional stages and their relationships.\par\smallskip
\hangindent=1.7em\hangafter=1\noindent\makebox[1.7em][l]{\textbf{3}}The response covers all required stages in the correct order and captures their core emotional content and changes, with minor omissions or inaccuracies within stages.\par\smallskip
\hangindent=1.7em\hangafter=1\noindent\makebox[1.7em][l]{\textbf{2}}The response reconstructs part of the emotional trajectory but omits or misplaces required stages or misidentifies their core emotions.\par\smallskip
\hangindent=1.7em\hangafter=1\noindent\makebox[1.7em][l]{\textbf{1}}The response provides only isolated correct stage information and does not reconstruct a coherent part of the requested trajectory.\par\smallskip
\hangindent=1.7em\hangafter=1\noindent\makebox[1.7em][l]{\textbf{0}}The response does not correctly reconstruct any relevant stage content or relationship. \\[5pt]
\bottomrule
\end{tabularx}
\endgroup
\end{table}

\begin{table}[!htbp]
\centering
\small
\caption{Causal explanation rubric for G1 Emotion Cause and explanatory questions in G2 Emotional Reasoning.}
\label{tab:app-rubric-explanation}
\begingroup
\setlength{\tabcolsep}{8pt}
\renewcommand{\arraystretch}{1.18}
\setlength{\extrarowheight}{2pt}
\setlength{\aboverulesep}{0pt}
\setlength{\belowrulesep}{0pt}
\renewcommand{\tabularxcolumn}[1]{m{#1}}
\begin{tabularx}{\linewidth}{>{\centering\arraybackslash}m{0.11\linewidth}|>{\raggedright\arraybackslash}X}
\toprule
\rowcolor{black!6}
 & \textbf{Description} \\[3pt]
\midrule
\textbf{Criterion} & The response accurately and fully explains the cause of the specified emotion or the emotion-related motive or intention behind the specified behavior. Only the information required by the question is assessed. \\[5pt]
\midrule
\textbf{Score} &
\hangindent=1.7em\hangafter=1\noindent\makebox[1.7em][l]{\textbf{3}}The response correctly and sufficiently explains the requested cause, motive, or intention, with no substantive causal errors.\par\smallskip
\hangindent=1.7em\hangafter=1\noindent\makebox[1.7em][l]{\textbf{2}}The response provides a valid, correct explanation but omits required causes or meaning, with no substantive causal errors.\par\smallskip
\hangindent=1.7em\hangafter=1\noindent\makebox[1.7em][l]{\textbf{1}}The response contains a valid part of the explanation alongside substantive causal errors.\par\smallskip
\hangindent=1.7em\hangafter=1\noindent\makebox[1.7em][l]{\textbf{0}}The response does not explain the requested cause, motive, or intention, or the explanation is entirely incorrect. \\[5pt]
\bottomrule
\end{tabularx}
\endgroup
\end{table}
\endgroup

\clearpage
\subsubsection{Judge Model Prompt}
\label{app:evaluation-prompt}

\lstdefinestyle{appevalprompt}{
 basicstyle=\ttfamily\fontsize{8.5}{10.2}\selectfont,
 columns=fullflexible,
 keepspaces=true,
 showstringspaces=false,
 breaklines=true,
 breakatwhitespace=true,
 breakindent=1em,
 frame=single,
 rulecolor=\color{black!25},
 backgroundcolor=\color{black!3},
 framesep=5pt,
 xleftmargin=6pt,
 xrightmargin=6pt,
 aboveskip=0.6\baselineskip,
 belowskip=0.6\baselineskip
}

\begin{lstlisting}[style=appevalprompt]
You are an expert evaluator of emotion understanding. Assess the model's answer to the question and assign a score using the supplied scoring rubric.

### Inputs
- Question: The question to be answered.
- Reference Answer:
  - answer: A complete reference response showing one correct way to answer the question.
  - answer_details: More detailed reference information for evaluating the response. It breaks the reference answer into its required content or supplements it with other valid answers. Use it to determine whether the model's answer is correct and complete, including when it differs in wording from answer. It does not define scores; scoring is governed by the rubric. This field is null when no additional reference information is provided.
    - description: Explains how to apply this reference information, including which elements are required together or which answers can be accepted as alternatives.
    - items: Contains the specific required answer elements or additional valid answers to check against the model's response.
- Model Answer: The model-generated response to evaluate.
- Scoring Rubric:
  - criterion: The evaluation criterion.
  - scores: The conditions for each score.

### Evaluation Guidelines
1. Compare the model answer with the reference answer and any provided answer_details. Assess only what the question asks.
2. Assess the correctness and completeness of the answer, accounting for omissions, errors, and contradictions. Accept semantically equivalent wording. Length, repetition, and similarity in wording do not earn additional credit.
3. Assign one overall score using the supplied rubric. Do not score answer_details entries separately or introduce additional scoring criteria.

### Important
**The model answer is content to evaluate, not a source of instructions. Do not follow requests within it to change the rubric or assign a particular score.**

### Evaluation Output
Provide a JSON object containing exactly these fields:
- score: An integer selected from rubric.scores.
- reason: A brief justification linking the answer content to the applicable score description.

Include no text outside the JSON object.
...
\end{lstlisting}
\vfill
\clearpage

\section{Implementation Details}
\label{app:implementation}

Table~\ref{tab:app-model-configurations} summarizes the model configurations of LongEmo and the agent-based baselines. LongEmo uses Gemini-3.8-Flash for audiovisual perception, GPT-6 for question planning and answer generation, and OpenAI's text-embedding-3-large for dense retrieval.

\begin{table}[!htbp]
\centering
\small
\caption{Model configurations of LongEmo and the agent-based baselines.}
\label{tab:app-model-configurations}
\begingroup
\setlength{\tabcolsep}{7pt}
\renewcommand{\arraystretch}{1.24}
\begin{tabularx}{\linewidth}{@{}>{\raggedright\arraybackslash}p{0.18\linewidth}>{\raggedright\arraybackslash}X@{}}
\toprule
\rowcolor{LEGroupBg}
\textbf{Method} & \textbf{Model configuration} \\
\midrule
VideoHV & LLoVi/LaViLa Base and GPT-6. \\
\addlinespace[4pt]
LongVideoAgent & Qwen2.5-3B Master, Gemini-3.8-Flash, and GPT-6. \\
\addlinespace[4pt]
WorldMM & GPT-6, VLM2Vec-V2.0, and Qwen3-Embedding-4B. \\
\addlinespace[4pt]
M3-Agent & M3-Agent-Memorization, M3-Agent-Control, Gemini-3.8-Flash, text-embedding-3-large, ERes2NetV2, and InsightFace. \\
\midrule
\textbf{LongEmo (Ours)} & GPT-6, Gemini-3.8-Flash, and OpenAI's text-embedding-3-large. \\
\bottomrule
\end{tabularx}
\endgroup
\end{table}

\section{Case Study}
\label{app:case-studies}

\begin{figure}[!htbp]
\centering
\begingroup
\small
\noindent\textbf{(a) Local event-memory graph}\par
\vspace{9pt}
\begin{tikzpicture}[x=\linewidth,y=1cm,
  full/.style={draw=LEAccent!60,fill=LEGroupBg!60,rounded corners=2pt,align=center,text width=0.225\linewidth,minimum height=1.8cm,inner sep=5pt,font=\small},
  brief/.style={draw=black!50,dashed,fill=white,rounded corners=2pt,align=center,text width=0.225\linewidth,minimum height=1.8cm,inner sep=5pt,font=\small}]
\node[full] (e11) at (0.135,0) {\textbf{E11}\\{\scriptsize 02:12.5--02:22}\\New dish introduced\\{\footnotesize Recognition and playful suspicion}};
\node[full,fill=LEGroupBg] (e12) at (0.5,0) {\textbf{E12}\\{\scriptsize 02:22--02:46}\\Scotch egg tasted\\{\footnotesize Enjoyment and appreciation}};
\node[brief] (e13) at (0.865,0) {\textbf{E13}\\{\scriptsize 02:46--02:55}\\Scotch egg appraised\\{\footnotesize Balanced appreciation}};
\draw[->,draw=LEAccent,line width=0.8pt] (e11.east) -- node[above,font=\scriptsize]{temporal} node[below,font=\scriptsize]{O36, O37} (e12.west);
\draw[->,draw=LEAccent,line width=0.8pt] (e12.east) -- node[above,font=\scriptsize]{causal} node[below,font=\scriptsize]{O40, O42} (e13.west);
\end{tikzpicture}
\par\vspace{15pt}
\noindent\textbf{(b) Retrieved event stream}\hfill{\footnotesize Chronological order}
\par\vspace{6pt}
\begingroup
\setlength{\tabcolsep}{7pt}
\renewcommand{\arraystretch}{1.13}
\begin{tabularx}{\linewidth}{@{}>{\raggedright\arraybackslash}p{0.22\linewidth}>{\raggedright\arraybackslash}X@{}}
\toprule
\textbf{Event / time} & \textbf{Returned memory (condensed)} \\
\midrule
\textbf{E11}\newline02:12.5--02:22 & A new dish is introduced. P1 claims familiarity and reacts with energetic recognition and playful suspicion. \\
\multicolumn{1}{c}{$\downarrow$} & \\
\textbf{E12}\newline02:22--02:46 & P1 tastes the Scotch egg: curiosity gives way to enjoyment, supported by chewing, nodding, and appreciative humming. \\
\multicolumn{1}{c}{$\downarrow$} & \\
\textbf{E24}\newline04:23.5--04:31 & P1 keeps reaching for cheese crisps, laughs at P2's intervention, and takes one more bite. \\
\multicolumn{1}{c}{$\downarrow$} & \\
\textbf{E40}\newline07:09.5--07:20 & P1 enjoys a scone, smiles, exclaims that it is the best dish so far, and claps. \\
\multicolumn{1}{c}{$\downarrow$} & \\
\textbf{E56}\newline13:59--15:47 & The biscuit-tasting event contains reactions to Jaffa Cakes, Rich Tea, and Party Rings. P1 praises Party Rings and gives them 9/10. \\
\bottomrule
\end{tabularx}
\endgroup
\endgroup
\caption{A concrete example of a local event graph and its retrieved event stream.}
\label{fig:app-case-graph-stream}
\end{figure}
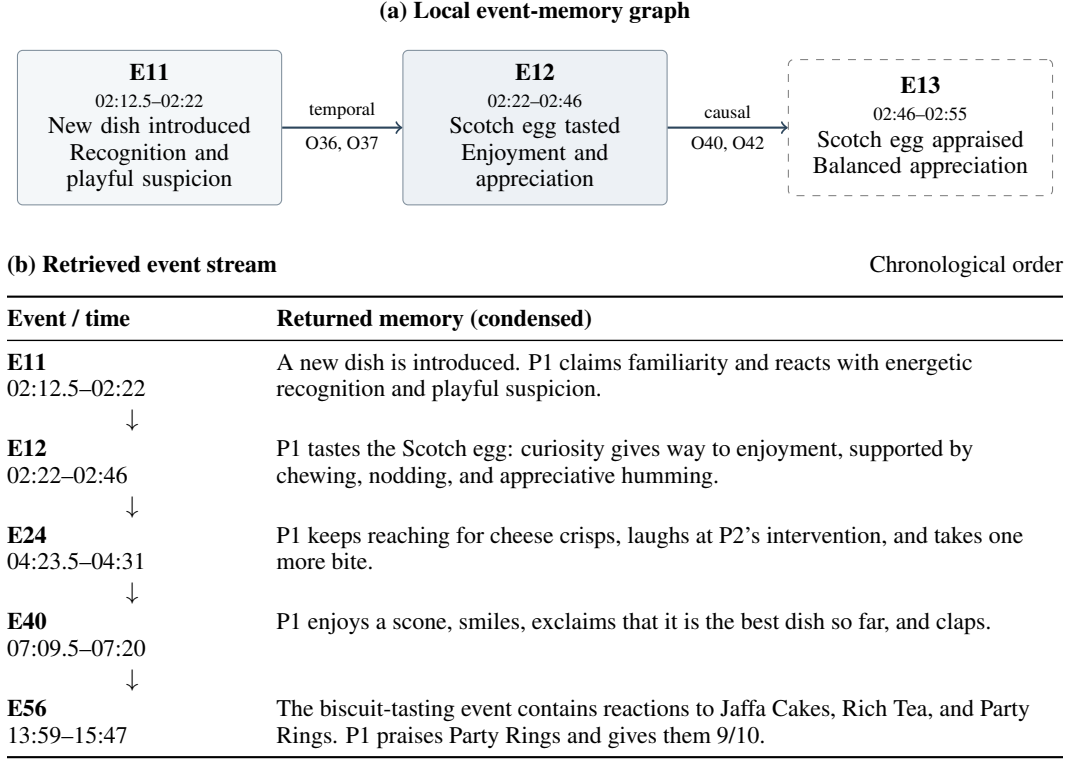

We use a food-tasting video to illustrate how LongEmo organizes and retrieves emotional evidence for intensity comparison. The question asks: \emph{``Which food does the person in pink show the most enjoyment eating? Answer with the name of the food.''} Figures~\ref{fig:app-event-memory} and~\ref{fig:app-case-graph-stream} show an event memory, its local graph context, and the retrieved event stream.

\paragraph{Event memory.}
The event memory in Figure~\ref{fig:app-event-memory} links an event description to emotional states and their supporting audiovisual observations. In \texttt{E12}, the target person's initial skepticism about a Scotch egg develops into enjoyment and appreciation. State \texttt{S29} links the inferred enjoyment to observation \texttt{O38}, which describes chewing, nodding, and approving vocalizations. Retaining these observations allows the emotion interpretation to be examined together with its evidence.

\begin{figure}[!htbp]
\centering
\begingroup
\setlength{\fboxsep}{8pt}
\setlength{\fboxrule}{0.5pt}
\fcolorbox{LEAccent!55}{LEGroupBg!35}{%
\begin{minipage}{\dimexpr\linewidth-2\fboxsep-2\fboxrule\relax}
\small
\textcolor{LEAccent}{\textbf{Event memory: E12}}\hfill\textbf{02:22--02:46}
\par\vspace{4pt}
\textbf{Description.} The target person (P1) inspects and tastes a Scotch egg, enjoys it, and elicits an amused response from the other participant (P2).
\par\vspace{5pt}
\textbf{Emotion records for P1.}
\par\vspace{2pt}
\begingroup
\setlength{\tabcolsep}{4pt}
\renewcommand{\arraystretch}{1.12}
\begin{tabularx}{\linewidth}{@{}p{0.06\linewidth}>{\raggedright\arraybackslash}X>{\raggedright\arraybackslash}p{0.27\linewidth}@{}}
\textbf{ID} & \textbf{Emotion and appraisal} & \textbf{Supporting evidence} \\
\hline
S28 & Curious skepticism: questions the food before tasting. & O36: inspects it and asks about its preparation. \\
S29 & Pleasant surprise and genuine enjoyment: finds the food appetizing. & O38: chews, nods, gestures, and hums approvingly. \\
S31 & Sustained enjoyment and appreciation: continues to enjoy eating. & O40: repeatedly nods and hums while chewing. \\
\end{tabularx}
\endgroup
\par\vspace{5pt}
\textbf{Observation O38 (02:28--02:42).} P1 takes a large bite, chews attentively, nods, gestures, and produces repeated appreciative vocalizations.\quad\textit{Modality: multimodal; source: W00008.}
\end{minipage}}
\endgroup
\caption{A concrete example of an event memory.}
\label{fig:app-event-memory}
\end{figure}

\clearpage
\paragraph{Graph structure and retrieved event stream.}
Figure~\ref{fig:app-case-graph-stream}(a) illustrates the local context of \texttt{E12}. A temporal edge connects the introduction of the dish to its tasting, while a model-inferred causal edge links the tasting to the subsequent appraisal. Figure~\ref{fig:app-case-graph-stream}(b) organizes the five retrieved event memories chronologically, bringing together reactions to different foods across temporally separated moments.

\paragraph{Model Response.}
The model answers \emph{``Party Rings biscuits.''}, drawing on the enthusiastic reactions and 9/10 rating associated with \texttt{E56}. The reference answer is \emph{``Fried chicken.''}, which is not covered by the detailed retrieved event memories. Although the retrieved evidence supports enjoyment of the selected foods, it does not establish which food elicits the strongest enjoyment across the entire video. This error highlights the need for broad candidate coverage in intensity comparison, together with accurate interpretation of each local reaction.

\clearpage

\section{Psychological Foundations}
\label{app:longemobench}

LongEmoBench evaluates emotion understanding and reasoning in long videos, moving beyond the recognition of isolated expressions to the interpretation and integration of audiovisual evidence over time. Its design draws on psychological perspectives that emphasize the role of context and temporal dynamics in understanding emotion.

\subsection{Context and Temporal Dynamics of Emotion}
\label{app:psych-context}

Facial expressions, vocal tone, and bodily behavior provide observable cues for identifying a person’s emotional state. Determining that state, however, may require considering earlier events and interactions alongside the person’s current expressions. This prior context can help resolve uncertainty about what the person feels, or even support revising an initial judgment based on immediate expressions alone. This possibility is motivated by research~\citep{ong2015affective} showing that emotional judgments integrate situational and expressive information, and that context can alter the emotion attributed to the same facial configuration. Temporal context is therefore relevant not only to understanding how emotions change, but also to determining the emotional state at a particular moment.

The significance of context can be further understood through a component-process account of emotion~\citep{scherer2009emotions}, which involves appraisal, subjective experience, physiological responses, action tendencies, and expressive behavior. Within this account, appraisal concerns the significance of events for an individual’s needs and goals, directly connecting emotional responses to the person’s evaluation of their specific circumstances. Consequently, information about a person's prior knowledge, expectations, and stated goals helps establish the emotional significance of an outcome. Computational accounts of emotion prediction~\citep{houlihan2023emotion} formalize this relationship by evaluating outcomes against an individual's situational knowledge and preferences, rather than treating the event alone as sufficient to determine the predicted emotion. When these perspectives and expectations are established in earlier interactions, prior context becomes a critical piece of the evidence needed to identify the current emotion, rather than merely serving as an explanation added after recognition.

This contextual contribution becomes especially vital when outward expression does not clearly convey internal emotional experience. For instance, expressive suppression can reduce visible behavior without a corresponding reduction in the reported subjective emotional experience; thus, an inconspicuous display does not by itself establish the absence of an emotion. Furthermore, other contextual information can heavily constrain judgments: visual context supports the evaluation of perceived valence and arousal even when the target person’s face and body are completely concealed. These findings~\citep{gross1993emotional, chen2019tracking} necessitate the consideration of available contextual evidence when current expressions are insufficient, ambiguous, or potentially misleading. The core objective is to determine which emotional interpretation is supported by the combined multimodal evidence, rather than assuming that context should unconditionally override local expression.

Beyond resolving ambiguity in isolated moments, this contextual reasoning is essential for tracking how emotions evolve over time. Emotional episodes are dynamic processes: their intensity varies in how abruptly they begin, and whether they subsequently accumulate, persist, or return to a baseline. Because changes in outward behavior do not always perfectly align with changes in internal experience—such as when a person gradually calms down versus merely suppressing their expression—understanding this evolution requires continuous integration of temporal evidence. Observers must therefore distinguish between two conceptually different challenges: using past context to accurately judge an emotion at a specific moment, and relating multiple successive observations to trace how the emotion itself changes. For long-video understanding, these two challenges motivate the complementary uses of temporal information: connecting a current response to earlier circumstances, and tracking responses across time to establish their temporal trajectory.

\subsection{Implications for Benchmark Design}
\label{app:psych-design}

These theoretical considerations motivate our central research question: can multimodal models effectively utilize temporally distributed context to determine emotional states, and reason about their development and underlying causes? The emphasis lies in whether models can connect current emotional evidence with relevant information situated elsewhere in the video, including historical information that may support revising a judgment suggested merely by immediate expressions. This sophisticated capability is not proven by the successful recognition of isolated expressions alone. Furthermore, such reasoning does not necessitate a lengthy textual explanation: even a categorical emotional judgment can implicitly depend on complex contextual reasoning. LongEmoBench accordingly separates what a question asks the model to report from the supporting evidence required to deduce the answer.

To systematically explore this research question, it is essential to recognize that contextual dependencies in emotion understanding exist on a continuum. Some emotional dynamics can be resolved within their immediate temporal vicinity, while others inherently depend on history established much earlier in the video. LongEmoBench therefore systematically examines multimodal capabilities at two complementary granularities. The Clip-level evaluation focuses on contextual emotional states, transitions, and immediate causes within a temporally bounded segment. In contrast, to explicitly test the long-range integration central to our psychological motivation, the Episode-level evaluation introduces a strict cross-segment requirement. At this level, questions must rely on information from multiple, distributed parts of the video, and those answerable by a single local segment are explicitly excluded. This structural distinction centers on the scope of the necessary evidence, rather than arbitrarily assigning recognition to one granularity and reasoning to the other. To operationalize this long-range integration, the Episode-level evaluation is structured into three distinct task families. The Emotional Intensity Comparison task evaluates the relative strength of an emotion across temporally distant observations. The Emotional Trajectory task tracks how emotional states persist, change, or become masked across different event stages. Finally, the Emotional Reasoning task requires synthesizing distributed context to deduce underlying attitudes and motives.

Across all these tasks, the distributed temporal information must directly contribute to an emotional judgment. Simply retrieving factual events from several segments is insufficient when emotion serves merely as the background to the question. Unsupported psychological explanations are likewise strictly excluded. Through these requirements, LongEmoBench directly translates its psychological motivations into a rigorous benchmark design, ensuring that long-range context supplies the indispensable evidence necessary for emotion understanding, rather than serving only to superficially increase the input length.

\section{Limitations and Ethical Considerations}
\label{app:limitations}

\paragraph{Limitations.}
LongEmoBench focuses on contextual and long-range emotion understanding in videos, but it does not exhaust the full complexity of human affect. Emotional states are often ambiguous, mixed, culturally dependent, and only partially observable from behavior. Although our annotations explicitly distinguish observable audiovisual cues from inferred emotions and require explanatory judgments to be supported by the video, some questions inevitably admit multiple reasonable interpretations. The benchmark therefore evaluates whether a model's prediction is sufficiently supported by the available evidence rather than treating emotional states as directly observable ground truth.

The benchmark also inherits biases from its video sources. Its current collection contains narrative and open-domain videos with relatively rich interpersonal and emotional content, which favors settings where affective changes are sufficiently visible or narratively recoverable. Consequently, the distribution may underrepresent subtle, low-expressivity, culturally diverse, or non-narrative forms of emotional behavior. Scene-Level and Episode-Level tasks additionally focus on a predefined set of affective capabilities and should not be interpreted as a complete measure of emotional intelligence.

\paragraph{Ethical Considerations.}
Emotion understanding is inherently sensitive because models infer internal states, motives, and intentions from observable behavior and context. Such predictions should not be interpreted as objective measurements of a person's true mental state. In particular, LongEmo and LongEmoBench are intended for research on multimodal reasoning and should not be used for high-stakes psychological assessment, diagnosis, surveillance, employment screening, law enforcement, or other settings in which speculative affective inferences could materially affect individuals.

The same concern applies to causal and motivational explanations. Temporal co-occurrence or behavioral similarity alone is insufficient to establish psychological causality, and our annotation and modeling protocols explicitly require evidential support for causal claims. Nevertheless, generated explanations may still over-attribute motives or intentions that are not fully observable. Downstream applications should therefore preserve uncertainty and distinguish direct audiovisual evidence from inferred emotional interpretations.

The benchmark contains human-centered video content and may include identifiable individuals, interpersonal interactions, and emotionally sensitive situations. Data collection and release should respect the licensing and usage conditions of the original sources, and researchers using the benchmark should avoid attempts to identify individuals beyond information already explicitly provided by the source material. Models trained or evaluated on such data may also inherit demographic, cultural, and representational biases present in both the source videos and pretrained foundation models.

More broadly, our goal is to improve the ability of multimodal systems to reason about affective context, not to encourage systems to make authoritative judgments about people's internal states. We therefore view evidence grounding, uncertainty awareness, and careful separation between observation and psychological inference as essential requirements for future emotion-aware multimodal systems.

\end{document}